\documentclass{article} 
\usepackage{iclr2025_conference,times}

\usepackage[utf8]{inputenc}
\usepackage[T1]{fontenc}

\usepackage{color,xcolor}
\usepackage{epsfig}
\usepackage{graphicx}

\usepackage{adjustbox}
\usepackage{array}
\usepackage{booktabs}
\usepackage{colortbl}
\usepackage{float,wrapfig}
\usepackage{hhline}
\usepackage{multirow}
\usepackage{subcaption}
\usepackage[font=small]{caption}
\usepackage{makecell}

\usepackage{amsmath,amsfonts,amsthm,amssymb}
\usepackage{bm}
\usepackage{nicefrac}
\usepackage{microtype}

\usepackage{changepage}
\usepackage{extramarks}
\usepackage{fancyhdr}
\usepackage{lastpage}
\usepackage{setspace}
\usepackage{soul}
\usepackage{xspace}
\usepackage{indentfirst}
\usepackage{pifont}
\usepackage{cuted}
\usepackage{wrapfig}
\definecolor{cvprblue}{rgb}{0.21,0.49,0.74}

\usepackage[pagebackref,breaklinks,colorlinks,citecolor=cvprblue]{hyperref}
\usepackage{url}

\usepackage{algorithm, algorithmic}
\usepackage{enumitem}

\usepackage{wasysym}
\usepackage{duckuments}
\usepackage{pifont}
\usepackage{duckuments}
\usepackage{fancyvrb}
\usepackage{fvextra}

\usepackage{amsmath,amsfonts,bm}

\def\eqref#1{equation~\ref{#1}}

\def\1{\bm{1}}

\def\vlambda{{\bm{\lambda}}}

\def\mL{{\bm{L}}}

\def\mQ{{\bm{Q}}}
\def\mR{{\bm{R}}}

\def\mU{{\bm{U}}}
\def\mV{{\bm{V}}}
\def\mW{{\bm{W}}}
\def\mX{{\bm{X}}}

\def\mSigma{{\bm{\Sigma}}}

\DeclareMathAlphabet{\mathsfit}{\encodingdefault}{\sfdefault}{m}{sl}
\SetMathAlphabet{\mathsfit}{bold}{\encodingdefault}{\sfdefault}{bx}{n}

\def\sR{{\mathbb{R}}}

\newcommand{\norm}[1]{\left\lVert#1\right\rVert}

\ifx\theorem\undefined
\newtheorem{theorem}{Theorem}[section]
\fi

\ifx\example\undefined
\newtheorem{example}{Example}[section]
\fi

\ifx\property\undefined
\newtheorem{property}{Property}
\fi

\ifx\lemma\undefined
\newtheorem{lemma}{Lemma}[section]
\fi

\ifx\proposition\undefined
\newtheorem{proposition}{Proposition}[section]
\fi

\ifx\remark\undefined
\newtheorem{remark}{Remark}
\fi

\ifx\corollary\undefined
\newtheorem{corollary}{Corollary}[section]
\fi

\ifx\definition\undefined
\newtheorem{definition}{Definition}[section]
\fi

\ifx\conjecture\undefined
\newtheorem{conjecture}{Conjecture}[section]
\fi

\ifx\axiom\undefined
\newtheorem{axiom}[theorem]{Axiom}
\fi

\ifx\claim\undefined
\newtheorem{claim}[theorem]{Claim}
\fi

\ifx\assumption\undefined
\newtheorem{assumption}{Assumption}
\fi

\ifx\problem\undefined
\newtheorem{problem}{Problem}[section]
\fi

\ifx\fact\undefined
\newtheorem{fact}{Fact}
\fi

\newcommand{\rbr}[1]{\left(#1\right)}
\newcommand{\sbr}[1]{\left[#1\right]}

\newcolumntype{L}[1]{>{\raggedright\let\newline\\\arraybackslash\hspace{0pt}}m{#1}}
\newcolumntype{C}[1]{>{\centering\let\newline\\\arraybackslash\hspace{0pt}}m{#1}}
\newcolumntype{R}[1]{>{\raggedleft\let\newline\\\arraybackslash\hspace{0pt}}m{#1}}

\newcommand{\sect}[1]{Section~\ref{sect:#1}}
\newcommand{\app}[1]{Appendix~\ref{app:#1}}

\newcommand{\eqn}[1]{Equation~\ref{eq:#1}}

\newcommand{\fig}[1]{Figure~\ref{fig:#1}}
\newcommand{\figs}[1]{Figures~\ref{fig:#1}}
\newcommand{\tbl}[1]{Table~\ref{tab:#1}}

\newcommand{\pps}[1]{Proposition~\ref{pps:#1}}
\newcommand{\fid}{Fr\'echet Inception Distance\xspace}
\newcommand{\lblfig}[1]{\label{fig:#1}}
\newcommand{\lblsect}[1]{\label{sect:#1}}
\newcommand{\lblapp}[1]{\label{app:#1}}
\newcommand{\lbleqn}[1]{\label{eq:#1}}
\newcommand{\lbltbl}[1]{\label{tab:#1}}

\newcommand{\lblpps}[1]{\label{pps:#1}}

\newcommand{\ignorethis}[1]{}

\def\naive{na\"{\i}ve\xspace}
\def\Naive{Na\"{\i}ve\xspace}
\def\naively{na\"{\i}vely\xspace}
\def\Naively{Na\"{\i}vely\xspace}

\makeatletter
\DeclareRobustCommand\onedot{\futurelet\@let@token\@onedot}
\def\@onedot{\ifx\@let@token.\else.\null\fi\xspace}

\def\ie{\emph{i.e}\onedot} 
 
 \def\versus{\emph{vs}\onedot}

\makeatother

\definecolor{citecolor}{rgb}{34,139,34}
\definecolor{mydarkblue}{rgb}{0,0.08,1}
\definecolor{mydarkgreen}{rgb}{0.02,0.6,0.02}
\definecolor{mydarkred}{rgb}{0.8,0.02,0.02}
\definecolor{mydarkorange}{rgb}{0.40,0.2,0.02}
\definecolor{mypurple}{RGB}{111,0,255}
\definecolor{myred}{rgb}{1.0,0.0,0.0}
\definecolor{mygold}{rgb}{0.75,0.6,0.12}
\definecolor{mydarkgray}{rgb}{0.66,0.66,0.66}

\newcommand{\myparagraph}[1]{\vspace{-2pt}\noindent\textbf{#1}}

\newcommand{\cmark}{\textcolor{green}{\ding{51}}}
\newcommand{\xmark}{\textcolor{red}{\ding{55}}}

\def\method{SVDQuant\xspace}
\def\engine{Nunchaku\xspace}

\definecolor{mitblue}{rgb}{0.88,0.95,0.96}

\title{
    \method: Absorbing Outliers by Low-Rank Components for 4-Bit Diffusion Models
}

\author{\textbf{Muyang Li}$^{1}$\thanks{Algorithm co-lead. \textsuperscript{\textdagger} System lead. $^\ddag$ Part of the work done during an internship at NVIDIA.} \space \footnotemark[3]
\quad
\textbf{Yujun Lin}$^{1}$\footnotemark[1]
\quad
\textbf{Zhekai Zhang}$^{1}$\footnotemark[2]
\quad
\textbf{Tianle Cai}$^{4}$
\quad
\textbf{Xiuyu Li}$^{5}$\footnotemark[3]\\
\textbf{Junxian Guo}$^{1,6}$
\quad
\textbf{Enze Xie}$^{2}$
\quad
\textbf{Chenlin Meng}$^{7}$ 
\quad
\textbf{Jun-Yan Zhu}$^{3}$
\quad
\textbf{Song Han}$^{1,2}$ \\
$^1$MIT
\quad
$^2$NVIDIA
\quad
$^3$CMU
\quad
$^4$Princeton
\quad
$^5$UC Berkeley
\quad
$^6$SJTU
\quad
$^7$Pika Labs \\
\url{https://hanlab.mit.edu/projects/svdquant}
}

\iclrfinalcopy 
\begin{document}

\maketitle

\begin{figure}[H]
    \vspace{-20pt}
    \centering    \includegraphics[width=\linewidth]{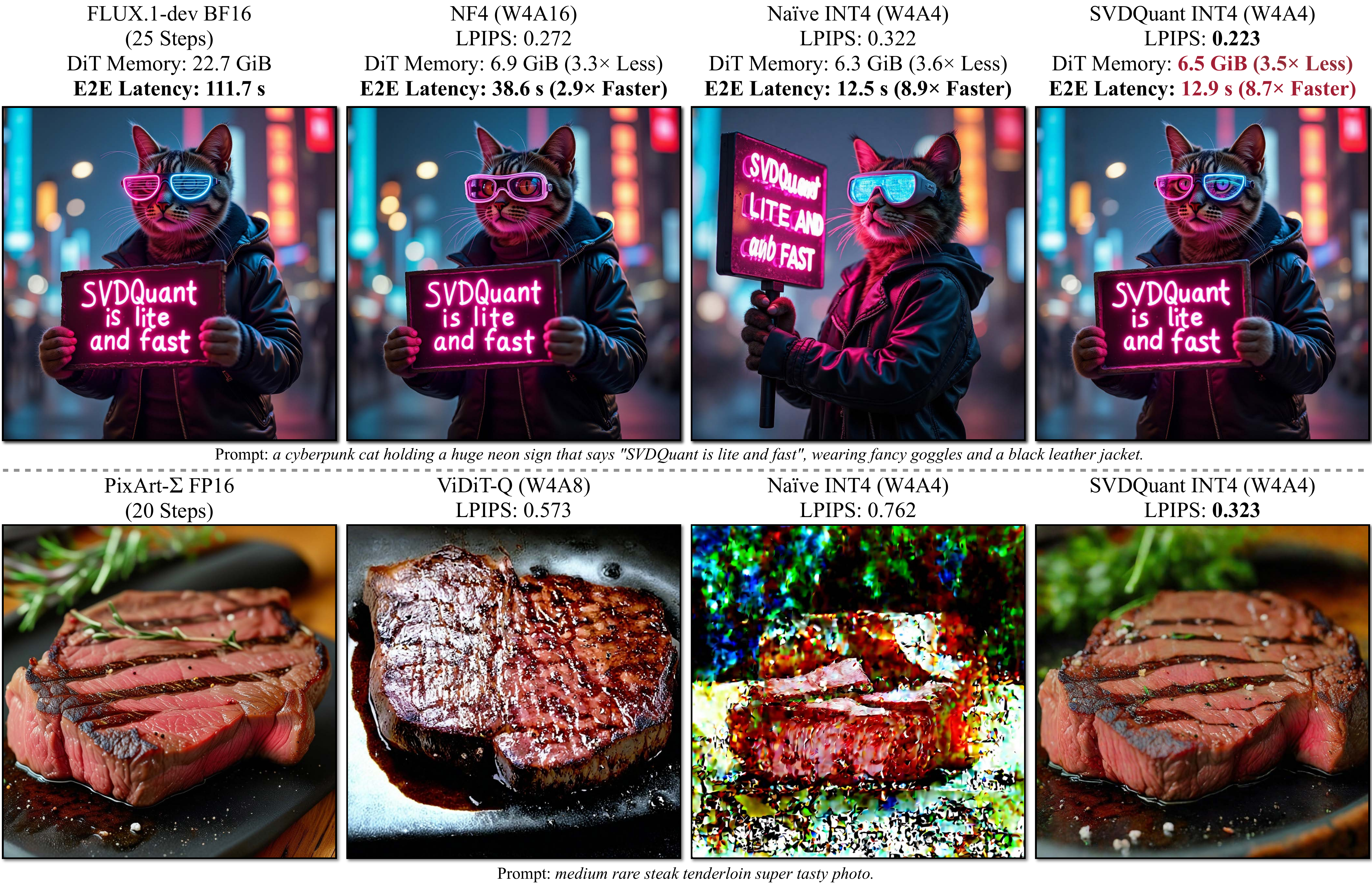}
    \vspace{-15pt}
    \caption{
    	\looseness=-1
        \method is a post-training quantization technique for 4-bit weights and activations that well maintains visual fidelity. On 12B FLUX.1-dev, it achieves 3.6× memory reduction compared to the BF16 model. By eliminating CPU offloading, it offers 8.7× speedup over the 16-bit model when on a 16GB laptop 4090 GPU, 3× faster than the NF4 W4A16 baseline. On PixArt-$\Sigma$, it demonstrates significantly superior visual quality over other W4A4 or even W4A8 baselines. ``E2E'' means the end-to-end latency including the text encoder and VAE decoder.
    }
    \vspace{-10pt}
    \lblfig{teaser}
\end{figure}
\begin{abstract}
	\looseness=-1
    Diffusion models can effectively generate high-quality images. However, as they scale,  rising memory demands and higher latency pose substantial deployment challenges. In this work, we aim to accelerate diffusion models by quantizing their weights and activations to 4 bits. At such an aggressive level, both weights and activations are highly sensitive, where existing post-training quantization methods like smoothing become insufficient. To overcome this limitation, we propose \textit{\method}, a new 4-bit quantization paradigm. Different from smoothing, which redistributes outliers between weights and activations, our approach \textit{absorbs} these outliers using a low-rank branch. We first consolidate the outliers by shifting them from activations to weights. Then, we use a high-precision, low-rank branch to take in the weight outliers with Singular Value Decomposition (SVD), while a low-bit quantized branch handles the residuals. This process eases the quantization on both sides. However, \naively running the low-rank branch independently incurs significant overhead due to extra data movement of activations, negating the quantization speedup. To address this, we co-design an inference engine \textit{\engine} that fuses the kernels of the low-rank branch into those of the low-bit branch to cut off redundant memory access. It can also seamlessly support off-the-shelf low-rank adapters (LoRAs) without re-quantization. Extensive experiments on SDXL, PixArt-$\Sigma$, and FLUX.1 validate the effectiveness of \method in preserving image quality. We reduce the memory usage for the 12B FLUX.1 models by 3.5×, achieving 3.0× speedup over the 4-bit weight-only quantization (W4A16) baseline on the 16GB laptop 4090 GPU with INT4 precision. On the latest RTX 5090 desktop with Blackwell architecture, we achieve a 3.1× speedup compared to the W4A16 model using NVFP4 precision. Our \href{https://github.com/mit-han-lab/deepcompressor}{quantization library}\footnote{Quantization library: \href{https://github.com/mit-han-lab/deepcompressor}{github.com/mit-han-lab/deepcompressor}} and \href{https://github.com/mit-han-lab/nunchaku}{inference engine}\footnote{Inference Engine: \href{https://github.com/mit-han-lab/nunchaku}{github.com/mit-han-lab/nunchaku}} are open-sourced.
\end{abstract}
\section{Introduction}

\looseness=-1
Diffusion models have shown remarkable capabilities in generating high-quality images~\citep{ho2020denoising}, with recent advances further enhancing user control over the generation process. Trained on vast data, these models can create stunning images from simple text prompts, unlocking diverse image editing and synthesis applications~\citep{meng2022sdedit,ruiz2023dreambooth,zhang2023adding}. 

\begin{wrapfigure}{r}{0.3\linewidth}
    \vspace{-15pt}
    \centering
    \includegraphics[width=\linewidth]{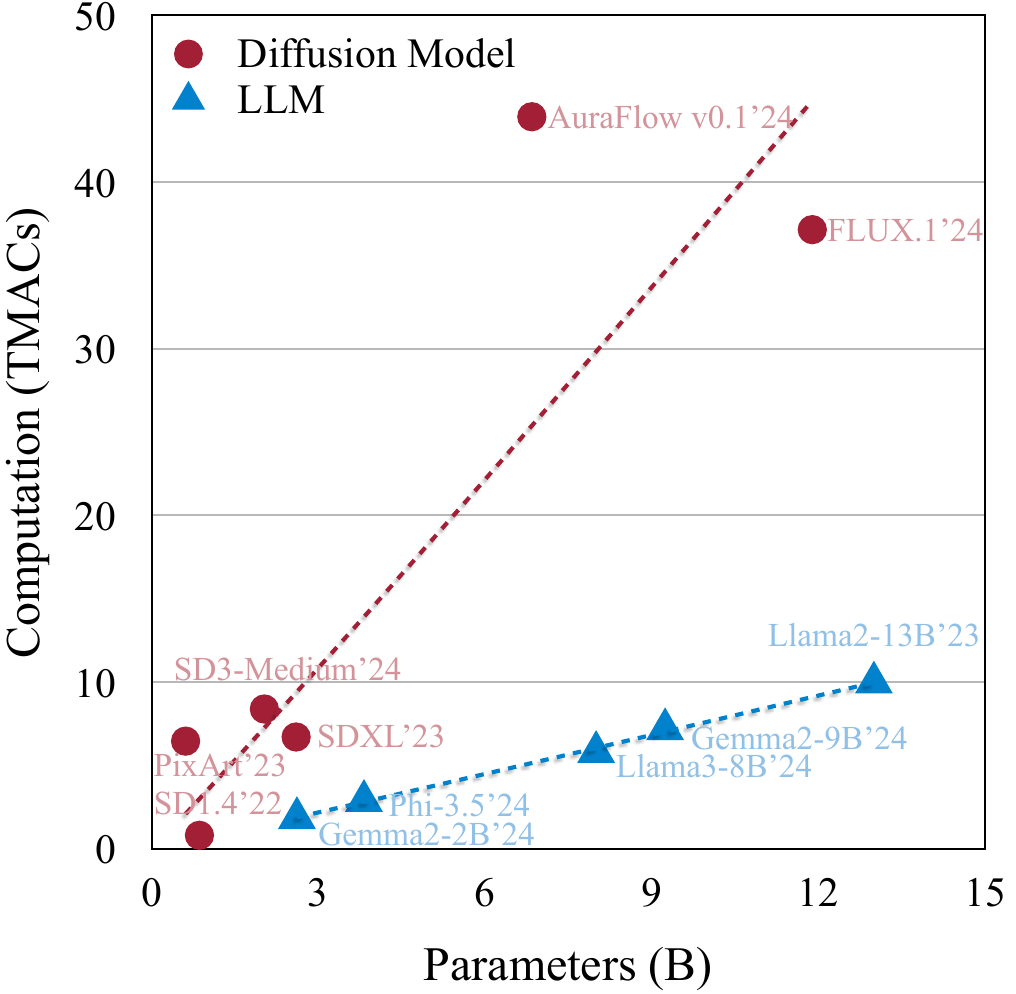}
    \vspace{-20pt}
    \caption{\looseness=-1 Computation \versus parameters for LLMs and diffusion models. LLMs' computation is measured with 512 context and 256 output tokens, and diffusion models' computation is for a single step. Dashed lines show trends.}
    \vspace{-15pt}
    \lblfig{trend}
\end{wrapfigure}

\looseness=-1
To pursue higher image quality and more precise text-to-image alignment, researchers are scaling up diffusion models. As shown in \fig{trend}, Stable Diffusion~(SD)~\citep{rombach2022high} 1.4 only has 800M parameters, while SDXL~\citep{podell2023sdxl} scales this up to 2.6B parameters. AuraFlow v0.1~\citep{auraflow0.1} extends this further to 6B parameters, with the latest model, FLUX.1~\citep{flux1}, pushing the boundary to 12B parameters. Compared to large language models (LLMs), diffusion models are significantly more computationally intensive. Their computational costs\footnote{Measured by the number of Multiply-Accumulate operations (MACs). 1 MAC=2 FLOPs.} increase more rapidly with model size, posing a prohibitive memory and latency barrier for real-world model deployment, particularly for interactive use cases that demand low latency.

\looseness=-1
As Moore's law slows down, hardware vendors are turning to low-precision inference to sustain performance improvements. For instance, NVIDIA's Blackwell Tensor Cores introduce a new 4-bit floating point (FP4) precision, doubling the performance compared to FP8~\citep{blackwell}. Therefore, using 4-bit inference to accelerate diffusion models is appealing. In the realm of LLMs, researchers have leveraged quantization to compress model sizes and boost inference speed~\citep{dettmers2022gpt3,xiao2023smoothquant}. However, unlike LLMs--where latency is primarily constrained by loading model weights on modern GPUs, especially with small batch sizes--diffusion models are heavily computationally bounded, even with a single batch. As a result, weight-only quantization cannot accelerate diffusion models. To achieve speedup on these devices, both weights and activations must be quantized to the same bit width; otherwise, the lower-precision weight will be upcast during computation, negating potential performance enhancements. 

In this work, we focus on quantizing both the weights and activations of diffusion models to 4 bits. This challenging and aggressive scheme is often prone to severe quality degradation. Existing methods like smoothing~\citep{xiao2023smoothquant,lin2024awq}, which transfer the outliers between the weights and activations, are less effective since both sides are highly vulnerable to outliers. To address this issue, we propose a general-purpose quantization paradigm, \textit{\method}. Our core idea is to use a low-cost branch to absorb outliers on both sides. To achieve this, as illustrated in \fig{intuition}, we first aggregate the outliers by migrating them from activation $\mX$ to weight $\mW$ via smoothing. Then we apply Singular Value Decomposition (SVD) to the updated weight, $\hat{\mW}$, decomposing it into a low-rank branch $\mL_1 \mL_2$ and a residual $\hat{\mW} - \mL_1 \mL_2$. The low-rank branch operates at 16 bits, allowing us to quantize only the residual to 4 bits, significantly reducing outliers and magnitude. However, naively running the low-rank branch separately incurs substantial memory access overhead, offsetting the speedup of 4-bit inference. To overcome this, we co-design a specialized inference engine \textit{\engine}, which fuses the low-rank branch computation into the 4-bit quantization and computation kernels. This design enables us to achieve measured inference speedup even with additional branches.

\method can quantize various text-to-image diffusion architectures into 4 bits, including both UNet~\citep{ho2020denoising,ronneberger2015u} and DiT~\citep{peebles2023scalable} backbones, while maintaining visual quality. It supports both INT4 and FP4 data types and integrates seamlessly with pre-trained low-rank adapters (LoRA)~\citep{hsu2022lowrank} without requiring re-quantization. To our knowledge, we are the first to successfully apply 4-bit post-training quantization to both the weights and activations of diffusion models, and achieve measured speedup on NVIDIA GPUs. On the latest 12B FLUX.1, our 4-bit models largely preserve the image quality and reduce the memory footprint of the original BF16 model by 3.5×. Furthermore, our INT4 and FP4 model delivers a 3.0× and 3.1× speedup over the NF4 weight-only quantized baseline on the 16GB laptop-level RTX 4090 and desktop-level RTX 5090 GPU, respectively. See \fig{teaser} for visual examples.

\begin{figure}[t]
    \centering
    \vspace{-15pt}
    \includegraphics[width=\linewidth]{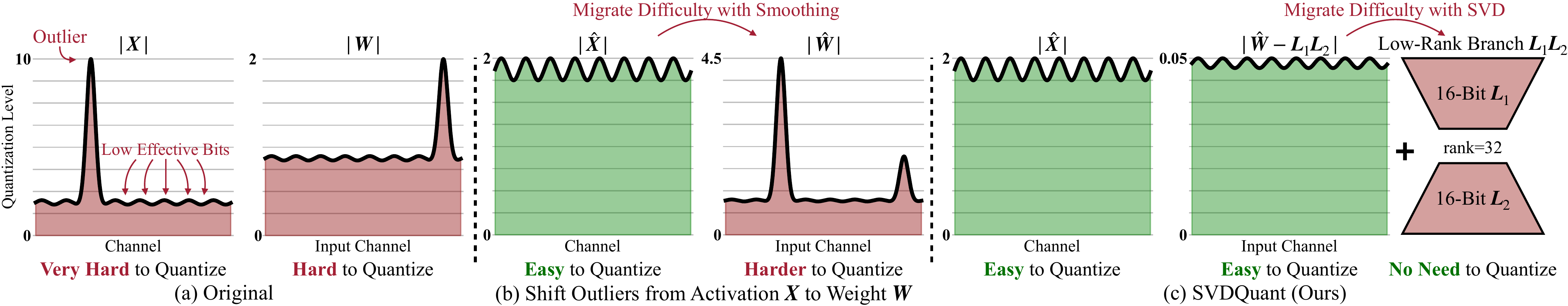}
    \vspace{-15pt}
	\caption{
        Overview of \method. (a) Originally, both the activation $\mX$ and weight $\mW$ contain outliers, making 4-bit quantization challenging. 
        (b) We migrate the outliers from the activation to weight, resulting in the updated activation $\hat{\mX}$ and weight $\hat{\mW}$. While $\hat{\mX}$ becomes easier to quantize, $\hat{\mW}$ now becomes more difficult.
        (c) \method further decomposes $\hat{\mW}$ into a low-rank component $\mL_1\mL_2$ and a residual $\hat{\mW}-\mL_1\mL_2$ with SVD. Thus, the quantization difficulty is alleviated by the low-rank branch, which runs at 16-bit precision. 
    }
    \lblfig{intuition}
    \vspace{-10pt}
\end{figure}

\section{Related Work}

\looseness=-1
\myparagraph{Diffusion models }~\citep{sohl2015deep,ho2020denoising} have emerged as a powerful class of generative models, known for generating high-quality samples by modeling the data distribution through an iterative denoising process. Recent advancements in text-to-image diffusion models~\citep{balaji2022ediffi,rombach2022high,podell2023sdxl} have already revolutionized content generation. Researchers further shifted from convolution-based UNet architectures~\citep{ronneberger2015u,ho2020denoising} to transformers~\citep{peebles2023scalable, bao2023all} and scaled up the model size~\citep{esser2024scaling}. However, diffusion models suffer from extremely slow inference speed due to their long denoising sequences and intense computation. To address this, various approaches have been proposed, including few-step samplers~\citep{zhang2022fast,zhang2022gddim,lu2022dpm} or distilling fewer-step models from pre-trained ones~\citep{salimans2021progressive,meng2022distillation,song2023consistency,luo2023latent,sauer2023adversarial,yin2024one,yin2024improved,kang2024distilling}. Another line of works choose to optimize or accelerate computation via efficient architecture design~\citep{li2023snapfusion,li2020gan,cai2024condition,liu2024linfusion}, quantization~\citep{shang2023post,li2023q}, sparse inference~\citep{li2022efficient,ma2024deepcache,ma2024learning}, and distributed inference~\citep{li2024distrifusion,wang2024pipefusion,chen2024asyncdiff}. This work focuses on quantizing the diffusion models to 4 bits to reduce the computation complexity. Our method can also be applied to few-step diffusion models to further reduce the latency (see \sect{Results}). 

\myparagraph{Quantization} has been recognized as an effective approach for LLMs to reduce the model size and accelerate inference~\citep{dettmers2022gpt3,frantar-gptq,xiao2023smoothquant,lin2025qserve,lin2024awq,kim2024squeezellm,zhao2024atom}. For diffusion models, Q-Diffusion~\citep{li2023q} and PTQ4DM~\citep{shang2023post} first achieved 8-bit quantization. Subsequent works refined these techniques with approaches like sensitivity analysis~\citep{yang2023efficient} and timestep-aware quantization~\citep{he2023ptqd,huang2024tfmq,liu2024enhanced,wang2024towards}. Some recent works extended the setting to text-to-image models~\citep{tang2023post,zhao2024mixdq}, DiT backbones~\citep{wu2024ptq4dit}, quantization-aware training~\citep{heefficientdm,zheng2024binarydm,wang2024quest,sui2024bitsfusion}, video generation~\citep{zhao2024vidit}, and different data types~\citep{liu2024hq}. Among these works, only MixDQ~\citep{zhao2024mixdq} and ViDiT-Q~\citep{zhao2024vidit} implement low-bit inference engines and report measured 8-bit speedup on GPUs. In this work, we push the boundary further by quantizing ffusion models to 4 bits, supporting both the integer or floating-point data types, compatible with the UNet backbone~\citep{ho2020denoising} and recent DiT architecture~\citep{peebles2023scalable}. Our co-designed inference engine, \engine, further ensures on-hardware speedup. Additionally, when applying LoRA to the model, existing methods require fusing the LoRA branch to the main branch and re-quantizing the model to avoid tremendous memory-access overhead in the LoRA branch. \engine cuts off this overhead via kernel fusion, allowing the low-rank branch to run efficiently as a separate branch, eliminating the need for re-quantization.

\looseness=-1
\myparagraph{Low-rank decomposition} has gained significant attention in deep learning for enhancing computational and memory efficiency~\citep{lora, zhao2024galore, jaiswal2024welore}. While directly applying this approach to model weights can reduce the compute and memory demands~\citep{hsu2022lowrank, yuan2023asvd, li2023losparse}, it often leads to performance degradation. Instead, \cite{yao2023zeroquant} combined it with quantization for model compression, employing a low-rank branch to compensate for the quantization error. Low-Rank Adaptation (LoRA)~\citep{lora} introduces another active line of research using low-rank matrices to adjust a subset of pre-trained weights for efficient fine-tuning.
This has sparked numerous advancements~\citep{dettmers2023qlora,guo2024lq, li2024loftq, xu2024qalora, meng2024pissa}, which combines quantized models with low-rank adapters to reduce memory usage during model fine-tuning. However, our work differs in two major aspects. First, our goal is different, as we aim to accelerate model inference through quantization, while previous works focus on model compression or efficient fine-tuning. Thus, they primarily consider weight-only quantization, resulting in no speedup. Second, as shown in our experiments (\fig{svd_fusion} and ablation study in \sect{Results}), directly applying these methods not only degrades the image quality, but also introduces significant overhead. In contrast, our method yields much better performance due to our joint quantization of weights and activations and overhead reduction of our inference engine \engine.
\section{Quantization Preliminary}
Quantization is an effective approach to accelerate linear layers in networks. Given a tensor $\mX$, the quantization process is defined as:
\begin{equation}\lbleqn{quantization_def}
    \mQ_\mX = \text{round}\left(\frac{\mX}{s_{\mX}}\right), s_{\mX} = \frac{\max (|\mX|)}{q_{\max}}.
\end{equation}
Here, $\mQ_\mX$ is the low-bit representation of $\mX$, $s_{\mX}$ is the scaling factor, and $q_{\max}$ is the maximum quantized value. For signed $k$-bit integer quantization, $q_{\max} = 2^{k-1}-1$. For 4-bit floating-point quantization with 1-bit mantissa and 2-bit exponent, $q_{\max} = 6$. Thus, the dequantized tensor can be formulated as $Q(\mX) = s_{\mX} \cdot \mQ_\mX$. For a linear layer with input $\mX$ and weight $\mW$, its computation can be approximated by
\begin{equation}
    \mX \mW \approx Q(\mX)Q(\mW) = s_{\mX}s_{\mW} \cdot \mQ_\mX \mQ_\mW.
\end{equation}
The same approximation applies to convolutional layers. To speed up computation, modern commercial GPUs require both $\mQ_\mX$ and $\mQ_\mW$ using the same bit width. Otherwise, the low-bit weights need to be upcast to match the higher bit width of activations, or vice versa, negating the speed advantage. Following the notation in QServe~\citep{lin2025qserve}, we denote $x$-bit weight, $y$-bit activation as W$x$A$y$. ``INT'' and ``FP'' refer to the integer and floating-point data types, respectively.

In this work, we focus on W4A4 quantization for acceleration, where outliers in both weights and activations place substantial obstacles.
Traditional methods to suppress these outliers include quantization-aware training (QAT)~\citep{heefficientdm} and  rotation~\citep{ashkboos2024quarot,liu2024spinquant,lin2025qserve}. QAT requires massive computing resources, especially for tuning models with more than 10B parameters such as FLUX.1. Rotation is inapplicable due to the usage of adaptive normalization layers~\citep{peebles2023scalable} in diffusion models. The runtime-generated normalization weights preclude the offline rotation with the weights of projection layers, while online rotation of both activations and weights incurs significant runtime overhead. 

\section{Method}
\lblsect{Method}

\looseness=-1
In this section, we first formulate our problem and discuss where the quantization error comes from.
Next, we present \method, a new W4A4 quantization paradigm for diffusion models. Our key idea is to introduce an additional low-rank branch that can absorb quantization difficulties in both weights and activations.
Finally, we provide a co-designed inference engine \engine with kernel fusion to minimize the overhead of the low-rank branches in the 4-bit model.

\begin{figure}[t]
    \centering
    \vspace{-15pt}
    \includegraphics[width=\linewidth]{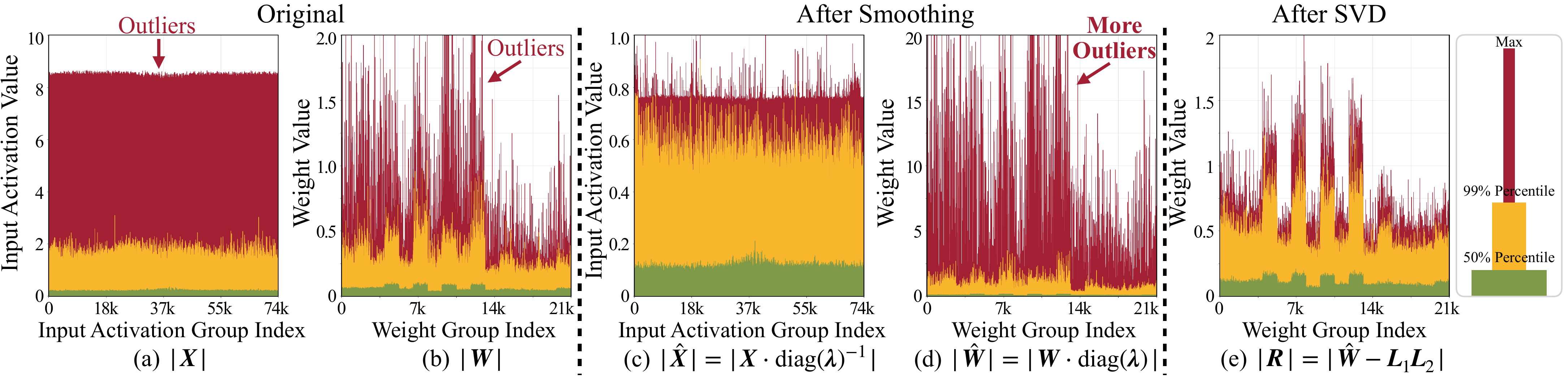}
    \vspace{-15pt}
	\caption{ 
        Example value distribution of inputs and weights in PixArt-$\Sigma$~\citep{chen2024pixart} $\vlambda$ is the smooth factor. Red indicates the outliers. Initially, both the input $\mX$ and weight $\mW$ contain significant outliers. After smoothing, the range of $\hat{\mX}$ is reduced with much fewer outliers, while $\hat{\mW}$ shows more outliers. Once the SVD low-rank branch $\mL_1\mL_2$ is subtracted, the residual $\mR$ has a narrower range and is free from outliers.
    }
    \lblfig{distribution}
    \vspace{-10pt}
\end{figure}

\subsection{Problem Formulation}
\looseness=-1
Consider a linear layer with input $\mX \in \sR^{b\times m}$ and weight $\mW \in \sR^{m\times n}$, where $b$ represents the batch size, and $m$ and $n$ denote the input and output channels, respectively. The quantization error can be defined as
\begin{equation} \lbleqn{error_def}
    E(\mX, \mW) = \norm{\mX \mW - Q(\mX)Q(\mW)}_F,
\end{equation}
where $\|\cdot\|_F$ denotes Frobenius Norm.
\begin{proposition}[Error decomposition] \lblpps{decomp}
The quantization error can be decomposed as follows:
\begin{align}
    \lbleqn{err_decomp}
    E(\mX, \mW) \le \norm{\mX}_F\norm{\mW - Q(\mW)}_F + \norm{\mX - Q(\mX)}_F(\norm{\mW}_F+\norm{\mW-Q(\mW)}_F).
\end{align}
\end{proposition}
See \app{proof1} for the proof. From the proposition, we can see that the error is bounded by four elements -- the magnitude of the weight and input, $\norm{\mW}_F$ and $\norm{\mX}_F$, and their respective quantization errors, $\norm{\mW-Q(\mW)}_F$ and $\norm{\mX-Q(\mX)}_F$. To minimize the overall quantization error, we aim to optimize these four terms.

\subsection{\method: Absorbing Outliers via Low-Rank Branch}
\lblsect{SVDQuant}
\myparagraph{Migrate outliers from activation to weight.}
Smoothing~\citep{xiao2023smoothquant,lin2024awq} is an effective approach for reducing outliers. We can smooth outliers in activations by scaling down the input $\mX$ and adjusting the weight matrix $\mW$ correspondingly using a per-channel smoothing factor $\vlambda \in \sR^m$. As shown in \fig{distribution}(a)(c), the smoothed input $\hat{\mX} = \mX \cdot \text{diag}(\vlambda)^{-1}$ exhibits reduced magnitude and fewer outliers, resulting in lower input quantization error. However, in \fig{distribution}(b)(d), the transformed weight $\hat{\mW} = \mW \cdot \text{diag}(\vlambda)$ has a significant increase in both magnitude and the presence of outliers, which in turn raises the weight quantization error. Consequently, the overall error reduction is limited.

\myparagraph{Absorb magnified weight outliers with a low-rank branch.} Our core insight is to introduce a 16-bit low-rank branch to further migrate the weight quantization difficulty. Specifically, we decompose the transformed weight as $\hat{\mW} = \mL_1 \mL_2 + \mR$, where $\mL_1 \in \sR^{m \times r}$ and $\mL_2 \in \sR^{r\times n}$ are two low-rank factors of rank $r$, and $\mR$ is the residual. Then $\mX \mW$ can be approximated as 
\begin{equation}
    \lbleqn{low-rank-compute}
    \mX \mW = \hat{\mX} \hat{\mW} = \hat{\mX} \mL_1\mL_2 + \hat{\mX} \mR \approx \underbrace{\hat{\mX} \mL_1\mL_2}_{\text{16-bit low-rank branch}} + \underbrace{Q(\hat{\mX}) Q(\mR)}_{\text{4-bit residual}}.
\end{equation}
Compared to direct 4-bit quantization, i.e., $Q(\hat{\mX}) Q(\mW)$, our method first computes the low-rank branch $\hat{\mX} \mL_1 \mL_2$ in 16-bit precision, and then approximates the residual $\hat{\mX} \mR$ with 4-bit quantization. Empirically, $r \ll \min(m, n)$, and is typically set to 16 or 32. As a result, the additional parameters and computation for the low-rank branch are negligible, contributing only $\frac{mr + nr}{mn}$ to the overall costs. However, it still requires careful system design to eliminate redundant memory access, which we will discuss in \sect{engine}.

From \eqn{low-rank-compute}, the quantization error can be expressed as
\begin{equation}
    \norm{\hat{\mX} \hat{\mW}-(\hat{\mX}\mL_1\mL_2+Q(\hat{\mX})Q(\mR))}_F = \norm{\hat{\mX} \mR - Q(\hat{\mX})Q(\mR)}_F = E(\hat{\mX}, \mR), 
\end{equation}
where $\mR = \hat{\mW}-\mL_1\mL_2$. According to \pps{decomp}, since $\hat{\mX}$ is already free from outliers, we only need to focus on optimizing the magnitude of $\mR$, $\norm{\mR}_F$ and its quantization error, $\norm{\mR - Q(\mR)}_F$.

\begin{proposition}[Quantization error bound]\lblpps{quant}
For any tensor $\mR$ and quantization method described in \eqn{quantization_def} as $Q(\mR) = s_\mR \cdot \mQ_\mR $. Assuming the elements of $\mR$ follow a distribution that satisfies the following regularity condition: There exists a constant $c$ such that
\begin{align}
    \mathbb{E} \sbr{\max(|\mR|)} \le c\cdot\mathbb{E} \sbr{\norm{\mR}_F} \lbleqn{regularity}.
\end{align}
Then, we have
\begin{align}
    \mathbb{E} \sbr{\norm{\mR - Q(\mR)}_F} \le \frac{c\sqrt{\text{size}(\mR)}}{q_{\max}} \cdot \mathbb{E}\sbr{\norm{\mR}_F}
\end{align}
where $\text{size}(\mR)$ denotes the number of elements in $\mR$. Especially if the elements of $\mR$ follow a normal distribution, \eqn{regularity} holds for $c = \sqrt{\frac{\log \rbr{\text{size}(\mR)}\pi}{\text{size}(\mR)}}$.
\end{proposition}
\begin{wrapfigure}{r}{0.3\linewidth}
    \vspace{-30pt}
    \centering
    \includegraphics[width=\linewidth]{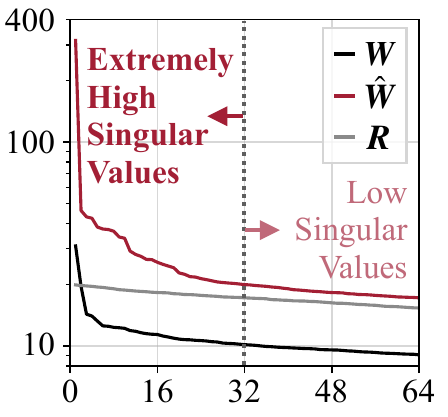}
    \vspace{-20pt}
    \caption{First 64 singular values of $\mW$, $\hat{\mW}$, and $\mR$. The first 32 singular values of $\hat{\mW}$ exhibit a steep drop, while the remaining values are much more gradual.}
    \vspace{-25pt}
    \lblfig{singular_values}
\end{wrapfigure}
See \app{proof2} for the proof. From this proposition, we obtain the intuition that the quantization error $\norm{\mR-Q(\mR)}_F$ is bounded by the magnitude of the residual $\norm{\mR}_F$. Thus, our goal is to find the optimal $\mL_1\mL_2$ that minimizes $\norm{\mR}_F = \norm{\hat{\mW} - \mL_1\mL_2}_F$, which can be solved by Singular Value Decomposition (SVD)~\citep{eckart1936approximation,mirsky1960symmetric}. Given the SVD of $\hat{\mW}=\mU\mSigma\mV$, the optimal solution is $\mL_1 = \mU\mSigma_{:,:r}$ and $\mL_2 = \mV_{:r,:}$. \fig{singular_values} illustrates the singular value distribution of the original weight $\mW$, transformed weight $\hat{\mW}$ and residual $\mR$. The singular values of the original weight $\mW$ are highly imbalanced. After smoothing, the singular value distribution of $\hat{\mW}$ becomes even sharper, with only the first several values being significantly larger. By removing these dominant values, the magnitude of the residual 
$\mR$ is dramatically reduced, as $\norm{\mR}_F = \sqrt{\sum_{i=r+1}^{\min(m,n)} \sigma_i^2}$, compared to the original magnitude $\norm{\hat{\mW}}_F = \sqrt{\sum_{i=1}^{\min(m,n)} \sigma_i^2}$, where $\sigma_i$ is the $i$-th singular value of $\hat{\mW}$. Furthermore, \fig{distribution}(d)(e) show that $\mR$ exhibits fewer outliers with a substantially compressed value range compared to $\hat{\mW}$. In practice, we further reduce quantization errors by iteratively updating the low-rank branch through decomposing $\mW - Q(\mR)$ and adjusting $\mR$ accordingly for several iterations, and then picking the result with the smallest error. 
\begin{figure}[t]
    \centering
    \vspace{-15pt}
    \includegraphics[width=\linewidth]{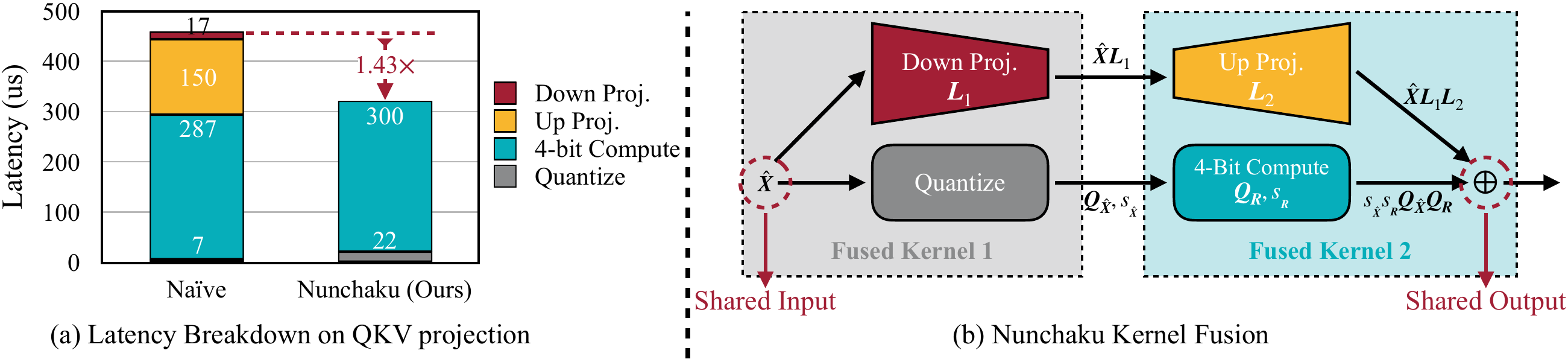}
    \vspace{-15pt}
	\caption{
		(a) \Naively running low-rank branch with rank 32 will introduce 57\% latency overhead due to extra read of 16-bit inputs in \textit{Down Projection} and extra write of 16-bit outputs in \textit{Up Projection}. Our \engine engine optimizes this overhead with kernel fusion. (b) \textit{Down Projection} and \textit{Quantize} kernels use the same input, while \textit{Up Projection} and \textit{4-Bit Compute} kernels share the same output. To reduce data movement overhead, we fuse the first two and the latter two kernels together.
    }
    \vspace{-10pt}
    \lblfig{svd_fusion}
\end{figure}

\subsection{\engine: Fusing Low-Rank and Low-Bit Branch Kernels}
\lblsect{engine}
Although the low-rank branch introduces negligible computation in theory, running it as a separate branch incurs large latency overhead—approximately 50\% of the 4-bit branch latency, as shown in \fig{svd_fusion}(a).
This occurs because, for a small rank $r$, even though the computational cost decreases significantly, the input and output activation sizes remain unchanged, shifting the bottleneck from computation to memory access. This issue worsens, especially when the activation cannot fit into the GPU L2 cache. For example, in the diffusion transformer block, the up projection in the low-rank branch for QKV projection is much slower since its output exceeds the available L2 cache, resulting in the extra DRAM load and store operations. Fortunately, the down projection $\mL_1$ in the low-rank branch shares the same input as the quantization kernel in the low-bit branch, while the up projection $\mL_2$ shares the same output as the 4-bit computation kernel, as illustrated in \fig{svd_fusion}(b). By fusing the down projection with the quantization kernel and the up projection with the 4-bit computation kernel, the low-rank branch can share the activations with the low-bit branch, eliminating the extra memory access and halving the number of kernel calls. As a result, our low-rank branch adds only 5$\sim$10\% latency, making it nearly cost-free. 
\section{Experiments}

\subsection{Setups}
\myparagraph{Models.} We benchmark our methods using FLUX.1~\citep{flux1}, PixArt-$\Sigma$~\citep{chen2024pixart}, SANA~\citep{xie2024sana}, Stable Diffusion XL (SDXL)~\citep{podell2023sdxl} and SDXL-Turbo~\citep{sauer2023adversarial}, including both the UNet~\citep{ronneberger2015u,ho2020denoising} and DiT~\citep{peebles2023scalable} backbones. See \app{Benchmark Models} for more details.

\myparagraph{Datasets.} Following previous works~\citep{li2023q,zhao2024mixdq,zhao2024vidit}, we randomly sample the prompts in COCO Captions 2024~\citep{chen2015microsoft} for calibration. To evaluate the generalization capability of our method, we sample 5K prompts from the MJHQ-30K~\citep{li2024playground} and the summarized Densely Captioned Images (sDCI)~\citep{urbanek2024picture} for benchmarking. See \app{Benchmark Datasets} for more details.

\looseness=-1
\myparagraph{Baselines.}
We compare \method against the following post-training quantization (PTQ) methods:
\begin{itemize}[leftmargin=*]
    \vspace{-5pt}
    \item \looseness=-1 4-bit NormalFloat (NF4) is an information-theoretically optimal 4-bit data type for weight-only quantization~\citep{dettmers2023qlora}, which assumes that weights follow a normal distribution.  We use the community-quantized NF4 FLUX.1 models~\citep{Lllyasviel} as the baselines.
    \item \looseness=-1 ViDiT-Q~\citep{zhao2024vidit} uses per-token quantization and smoothing~\citep{xiao2023smoothquant} to alleviate the outliers across different batches and tokens and achieves lossless 8-bit quantization on PixArt-$\Sigma$.
    \item MixDQ~\citep{zhao2024mixdq} identifies the outliers in the begin-of-sentence token of text embedding and protects them with 16-bit pre-computation. This method enables up to W4A8 quantization with negligible performance degradation on SDXL-Turbo.
    \item \href{https://developer.nvidia.com/blog/tensorrt-accelerates-stable-diffusion-nearly-2x-faster-with-8-bit-post-training-quantization/}{TensorRT} contains an industry-level PTQ toolkit to quantize the diffusion models to 8 bits. It uses smoothing and only calibrates activations over a selected timestep range with a percentile scheme.
\end{itemize}

\myparagraph{Metrics.} Following previous works~\citep{li2022efficient,li2024distrifusion}, we evaluate image quality and image similarity with respect to the 16-bit models' results. For image quality assessment, we use \fid (FID, lower is better) to measure the distribution distance between the generated images and the ground-truth images~\citep{heusel2017gans,parmar2021cleanfid}. Besides, we employ Image Reward (higher is better) to approximate the human rating of the generated images~\citep{xu2024imagereward}. We use LPIPS (lower is better) to measure the perceptual similarity~\citep{zhang2018perceptual} and Peak Signal Noise Ratio (PSNR, higher is better) to measure the numerical similarity of the images from the 16-bit models. Please refer to our \app{Quality Results} for more metrics (CLIP IQA~\citep{wang2022exploring}, CLIP Score~\citep{hessel2021clipscore} and SSIM\footnote{\url{https://en.wikipedia.org/wiki/Structural_similarity_index_measure}}).

\renewcommand \arraystretch{1.}
\begin{table}[t]
    \setlength{\tabcolsep}{3.4pt}
    \vspace{-15pt}
    \caption{
        Quantitative quality comparisons across different models. RTN stands for round-to-nearest. IR means ImageReward. Our 8-bit results closely match the quality of the 16-bit models. Moreover, our 4-bit results outperform other 4-bit baselines, effectively preserving the visual quality of 16-bit models.
    }
    \vspace{-5pt}
    \lbltbl{quality}
    \scriptsize \centering
    \begin{tabular}{cccccccccccc}
    \toprule
    & & & & \multicolumn{4}{c}{MJHQ} & \multicolumn{4}{c}{sDCI} \\
    \cmidrule(lr){5-8} \cmidrule(lr){9-12}
    Backbone & Model & Precision & Method & \multicolumn{2}{c}{Quality} & \multicolumn{2}{c}{Similarity} & \multicolumn{2}{c}{Quality} & \multicolumn{2}{c}{Similarity} \\
    \cmidrule(lr){5-6} \cmidrule(lr){7-8} \cmidrule(lr){9-10} \cmidrule(lr){11-12}
    & & & & FID ($\downarrow$) & IR ($\uparrow$) & LPIPS ($\downarrow$) & PSNR( $\uparrow$) & FID ($\downarrow$) & IR ($\uparrow$) & LPIPS ($\downarrow$) & PSNR ($\uparrow$) \\
    
    \midrule
    \multirow{30}{*}{DiT} & \multirow{7}{*}{\makecell{FLUX.1\\-dev\\(50 Steps)}} & BF16 & -- & 20.3 & 0.953 & -- & -- & 24.8 & 1.02 & -- & -- \\
    \cmidrule{3-12}
    & & \cellcolor{mitblue}INT W8A8 & \cellcolor{mitblue}Ours & \cellcolor{mitblue}20.4 & \cellcolor{mitblue}0.948 & \cellcolor{mitblue}0.089 & \cellcolor{mitblue}27.0 & \cellcolor{mitblue}24.7 & \cellcolor{mitblue}1.02 & \cellcolor{mitblue}0.106 & \cellcolor{mitblue}24.9  \\
    \cmidrule{3-12}
    & & W4A16 & NF4 & 20.6 & 0.910 & 0.272 & 19.5 & 24.9 & 0.986 & 0.292 & 18.2 \\
    & & \cellcolor{mitblue}INT W4A4 & \cellcolor{mitblue}Ours & \cellcolor{mitblue}\textbf{19.9} & \cellcolor{mitblue}0.935 & \cellcolor{mitblue}0.223 & \cellcolor{mitblue}21.0 & \cellcolor{mitblue}\textbf{24.2} & \cellcolor{mitblue}\textbf{1.01} & \cellcolor{mitblue}0.240 & \cellcolor{mitblue}19.7 \\
    & & \cellcolor{mitblue}NVFP W4A4 & \cellcolor{mitblue}Ours & \cellcolor{mitblue}20.4 & \cellcolor{mitblue}\textbf{0.937} & \cellcolor{mitblue}\textbf{0.208} & \cellcolor{mitblue}\textbf{21.4} & \cellcolor{mitblue}24.7 & \cellcolor{mitblue}\textbf{1.01} & \cellcolor{mitblue}\textbf{0.218} & \cellcolor{mitblue}\textbf{20.2} \\
    
    \cmidrule{2-12}
    & \multirow{7}{*}{\makecell{FLUX.1\\-schnell\\(4 Steps)}} & BF16 & -- & 19.2 & 0.938 & -- & -- & 20.8 & 0.932 & -- & -- \\
    \cmidrule{3-12}
    & & \cellcolor{mitblue}INT W8A8 & \cellcolor{mitblue}Ours & \cellcolor{mitblue}19.2 & \cellcolor{mitblue}0.966 & \cellcolor{mitblue}0.120 & \cellcolor{mitblue}22.9 & \cellcolor{mitblue}20.7 & \cellcolor{mitblue}0.975 & \cellcolor{mitblue}0.133 & \cellcolor{mitblue}21.3 \\
    \cmidrule{3-12}
    & & W4A16 & NF4 & 18.9 & 0.943 & 0.257 & 18.2 & 20.7 & 0.953 & 0.263 & 17.1 \\
    & & \cellcolor{mitblue}INT W4A4 & \cellcolor{mitblue}Ours & \cellcolor{mitblue}\textbf{18.3} & \cellcolor{mitblue}0.951 & \cellcolor{mitblue}0.258 & \cellcolor{mitblue}18.3 & \cellcolor{mitblue}\textbf{20.1} & \cellcolor{mitblue}\textbf{0.979} & \cellcolor{mitblue}0.260 & \cellcolor{mitblue}17.2 \\
    & & \cellcolor{mitblue}NVFP W4A4 & \cellcolor{mitblue}Ours & \cellcolor{mitblue}19.0 & \cellcolor{mitblue}\textbf{0.968} & \cellcolor{mitblue}\textbf{0.227} & \cellcolor{mitblue}\textbf{19.0} & \cellcolor{mitblue}20.5 & \cellcolor{mitblue}\textbf{0.979} & \cellcolor{mitblue}\textbf{0.226} & \cellcolor{mitblue}\textbf{18.1} \\
    
    \cmidrule{2-12}
    & \multirow{8}{*}{\makecell{PixArt-$\Sigma$\\(20 Steps)}} & FP16 & -- & 16.6 & 0.944 & -- & -- & 24.8 & 0.966 \\
    \cmidrule{3-12}
    & & INT W8A8 & ViDiT-Q & \textbf{15.7} & 0.944 & 0.137 & 22.5 & \textbf{23.5} & \textbf{0.974} & 0.163 & 20.4 \\
    & & \cellcolor{mitblue}INT W8A8 & \cellcolor{mitblue}Ours & \cellcolor{mitblue}16.3 & \cellcolor{mitblue}\textbf{0.955} & \cellcolor{mitblue}\textbf{0.109} & \cellcolor{mitblue}\textbf{23.7} & \cellcolor{mitblue}24.2 & \cellcolor{mitblue}0.969 & \cellcolor{mitblue}\textbf{0.129} & \cellcolor{mitblue}\textbf{21.8} \\
    \cmidrule{3-12}
    & & INT W4A8 & ViDiT-Q & 37.3 & 0.573 & 0.611 & 12.0 & 40.6 & 0.600 & 0.629 & 11.2 \\
    & & INT W4A4 & ViDiT-Q & 412 & -2.27 & 0.854 & 6.44 & 425 & -2.28 & 0.838 & 6.70 \\
    & & \cellcolor{mitblue}INT W4A4 & \cellcolor{mitblue}Ours & \cellcolor{mitblue}19.2 & \cellcolor{mitblue}0.878 & \cellcolor{mitblue}0.323 & \cellcolor{mitblue}17.6 & \cellcolor{mitblue}25.9 & \cellcolor{mitblue}0.918 & \cellcolor{mitblue}0.352 & \cellcolor{mitblue}16.5 \\
    & & \cellcolor{mitblue}NVFP W4A4 & \cellcolor{mitblue}Ours & \cellcolor{mitblue}\textbf{16.6} & \cellcolor{mitblue}\textbf{0.940} & \cellcolor{mitblue}\textbf{0.271} & \cellcolor{mitblue}\textbf{18.5} & \cellcolor{mitblue}\textbf{22.9} & \cellcolor{mitblue}\textbf{0.971} & \cellcolor{mitblue}\textbf{0.298} & \cellcolor{mitblue}\textbf{17.2} \\
    
    \cmidrule{2-12}
    & & BF16 & -- & 20.6 & 0.952 & -- & -- & 29.9 & 0.847 & -- & -- \\
    \cmidrule{3-12}
    & SANA& INT W4A4 & RTN & 20.5 & 0.894 & 0.339 & 15.3 & 28.6 & 0.807 & 0.371 & 13.8 \\
    & -1.6B& \cellcolor{mitblue}INT W4A4 & \cellcolor{mitblue}Ours & \cellcolor{mitblue}\textbf{19.3} & \cellcolor{mitblue}0.935 & \cellcolor{mitblue}0.220 & \cellcolor{mitblue}17.8 & \cellcolor{mitblue}\textbf{28.1} & \cellcolor{mitblue}\textbf{0.846} & \cellcolor{mitblue}0.242 & \cellcolor{mitblue}16.2 \\
    & (20 Steps)& NVFP W4A4 & RTN & 19.7 & 0.932 & 0.237 & 17.3 & 29.0 & 0.829 & 0.265 & 15.6 \\
    & & \cellcolor{mitblue}NVFP W4A4 & \cellcolor{mitblue}Ours & \cellcolor{mitblue}20.0 & \cellcolor{mitblue}\textbf{0.955} & \cellcolor{mitblue}\textbf{0.177} & \cellcolor{mitblue}\textbf{19.0} & \cellcolor{mitblue}29.3 & \cellcolor{mitblue}\textbf{0.846} & \cellcolor{mitblue}\textbf{0.196} & \cellcolor{mitblue}\textbf{17.3} \\
    
    \midrule
    \multirow{15}{*}{UNet} & \multirow{9}{*}{\makecell{SDXL\\-Turbo\\(4 Steps)}} & FP16 & --& 24.3 & 0.845 & -- & -- & 24.7 & 0.705 & -- & --\\
    \cmidrule{3-12}
    & & INT W8A8 & MixDQ & \textbf{24.1} & 0.834 & 0.147 & 21.7 & 25.0 & 0.690 & 0.157 & 21.6 \\
    & & \cellcolor{mitblue}INT W8A8 & \cellcolor{mitblue}Ours & \cellcolor{mitblue}24.3 & \cellcolor{mitblue}\textbf{0.845} & \cellcolor{mitblue}\textbf{0.100} & \cellcolor{mitblue}\textbf{24.0} & \cellcolor{mitblue}\textbf{24.8} & \cellcolor{mitblue}\textbf{0.701} & \cellcolor{mitblue}\textbf{0.110} & \cellcolor{mitblue}\textbf{23.7} \\
    \cmidrule{3-12}
    & & INT W4A8 & MixDQ & 27.7 & 0.708 & 0.402 & 15.7 & 25.9 & 0.610 & 0.415 & 15.7 \\
    & & INT W4A4 & MixDQ & 353 & -2.26 & 0.685 & 11.0 & 373 & -2.28 & 0.686 & 11.3 \\
    & & \cellcolor{mitblue}INT W4A4 & \cellcolor{mitblue}Ours & \cellcolor{mitblue}24.6 & \cellcolor{mitblue}0.816 & \cellcolor{mitblue}0.262 & \cellcolor{mitblue}18.1 & \cellcolor{mitblue}26.0 & \cellcolor{mitblue}0.671 & \cellcolor{mitblue}0.272 & \cellcolor{mitblue}18.0 \\
    & & \cellcolor{mitblue}NVFP W4A4 & \cellcolor{mitblue}Ours & \cellcolor{mitblue}\textbf{24.4} & \cellcolor{mitblue}\textbf{0.832} & \cellcolor{mitblue}\textbf{0.231} & \cellcolor{mitblue}\textbf{18.9} & \cellcolor{mitblue}\textbf{25.2} & \cellcolor{mitblue}\textbf{0.688} & \cellcolor{mitblue}\textbf{0.238} & \cellcolor{mitblue}\textbf{18.9} \\
    \cmidrule{2-12}
    
    & \multirow{6}{*}{\makecell{SDXL\\(30 Steps)}} & FP16 & -- & 16.6 & 0.729 & -- & -- & 22.5 & 0.573 & -- & -- \\
    \cmidrule{3-12}
    & & INT W8A8 & TensorRT & 20.2 & 0.591 & 0.247 & 22.0 & 25.4 & 0.453 & 0.265 & 21.7 \\
    & & \cellcolor{mitblue}INT W8A8 & \cellcolor{mitblue}Ours & \cellcolor{mitblue}\textbf{16.6} & \cellcolor{mitblue}\textbf{0.718} & \cellcolor{mitblue}\textbf{0.119} & \cellcolor{mitblue}\textbf{26.4} & \cellcolor{mitblue}\textbf{22.4} & \cellcolor{mitblue}\textbf{0.574} & \cellcolor{mitblue}\textbf{0.129} & \cellcolor{mitblue}\textbf{25.9} \\
    \cmidrule{3-12}
    & & \cellcolor{mitblue}INT W4A4 & \cellcolor{mitblue}Ours & \cellcolor{mitblue}20.6 & \cellcolor{mitblue}0.601 & \cellcolor{mitblue}0.288 & \cellcolor{mitblue}21.0 & \cellcolor{mitblue}26.2 & \cellcolor{mitblue}0.477 & \cellcolor{mitblue}0.307 & \cellcolor{mitblue}20.7 \\
    & & \cellcolor{mitblue}NVFP W4A4 & \cellcolor{mitblue}Ours & \cellcolor{mitblue}\textbf{18.3} & \cellcolor{mitblue}\textbf{0.640} & \cellcolor{mitblue}\textbf{0.250} & \cellcolor{mitblue}\textbf{21.8} & \cellcolor{mitblue}\textbf{23.9} & \cellcolor{mitblue}\textbf{0.502} & \cellcolor{mitblue}\textbf{0.261} & \cellcolor{mitblue}\textbf{21.7} \\   
    \bottomrule
    \end{tabular}
    \vspace{-10pt}
\end{table}

\looseness=-1
\myparagraph{Implementation details.} Please refer to \app{Implementation Details} fore more details.

\subsection{Results}
\lblsect{Results}
\myparagraph{Visual quality results.} 
We report the quantitative results in \tbl{quality} across various models and precision levels, and show some corresponding 4-bit qualitative comparisons in \fig{visual}. Among all models, our 8-bit results can perfectly mirror the 16-bit results, achieving PSNR higher than 21, beating all other 8-bit baselines. On FLUX.1-dev, our INT8 PSNR even reaches 27 on MJHQ.

\looseness=-1
For 4-bit quantization, NVFP4 outperforms INT4, thanks to the native hardware support of smaller microscaling group size on Blackwell. On FLUX.1, our \method consistently surpasses the NF4 W4A16 baseline regarding all metrics. For the dev variant, our method even exceeds the original BF16 model regarding Image Reward, suggesting stronger human preference. On PixArt-$\Sigma$, while our INT4 method shows slight degradation, our NVFP4 model achieves a comparable score to the FP16 model. This is likely due to PixArt-$\Sigma$'s highly compact model size (600M parameters), which benefits from a smaller group size. Remarkably, our INT4 and NVFP4 models significantly outperform ViDiT-Q's W4A8 results by a large margin across all metrics. Note that our FP16 PixArt-$\Sigma$ model differs slightly from ViDiT's, though both offer the same quality. For fair comparisons, ViDiT-Q's similarity results are calculated using their FP16 results.

For UNet-based models, on SDXL-Turbo, our 4-bit models substantially outperform MixDQ W4A8, and our FID scores are on par with the FP16 models, indicating no quality loss. On SDXL, our INT4 and NVFP4 results achieve comparable quality to TensorRT's W8A8 performance, which represents the 8-bit SoTA. As shown in \fig{sdxl-appendix} in the Appendix, our visual quality only shows minor degradation.

\begin{figure}[!t]
\captionsetup{font=small}
\centering
\vspace{-15pt}
\includegraphics[width=\textwidth]{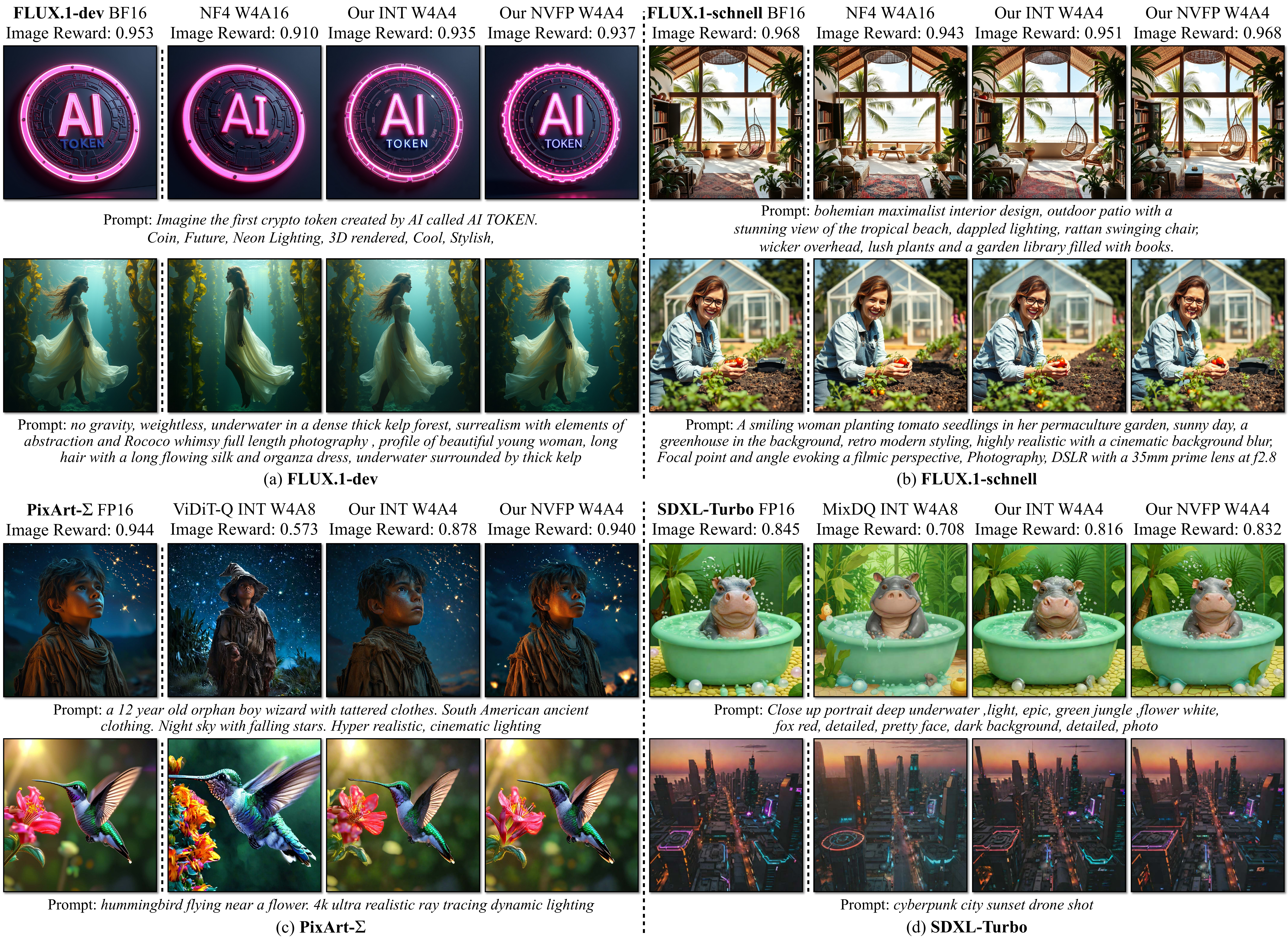}
\vspace{-15pt}
    \caption{\looseness=-1 Qualitative visual results on MJHQ. Image Reward is calculated over the entire dataset. On FLUX.1 models, our 4-bit models outperform the NF4 W4A16 baselines, demonstrating superior text alignment and closer similarity to the 16-bit models. For instance, NF4 misses the swinging chair in the top right example. On PixArt-$\Sigma$ and SDXL-Turbo, our 4-bit results demonstrate noticeably better visual quality than ViDiT-Q's and MixDQ's W4A8 results.}
\lblfig{visual}
 \vspace{-10pt}
\end{figure}

\begin{figure}[t]
    \centering
    \includegraphics[width=\linewidth]{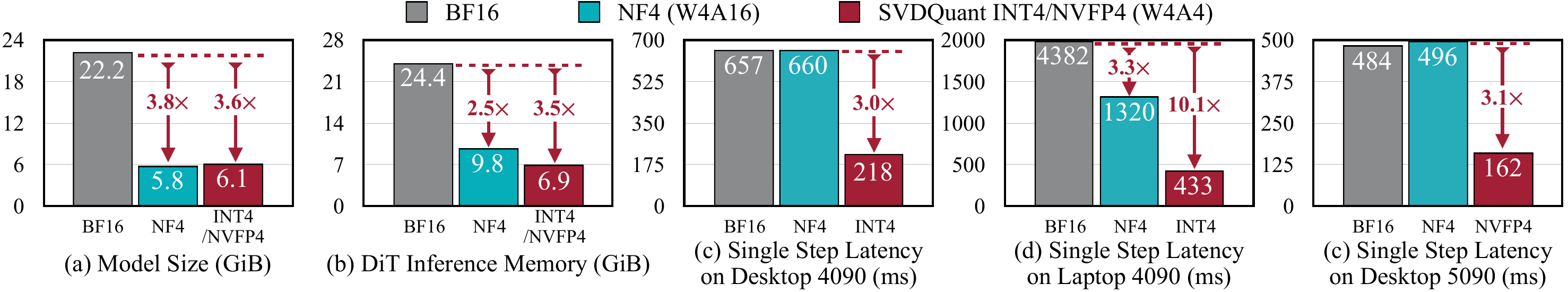}
    \vspace{-15pt}
	\caption{
        \method reduces the 12B FLUX.1 model size by 3.6× and cuts the 16-bit model's memory usage by 3.5×. With \engine, our INT4 model runs 3.0× faster than the NF4 W4A16 baseline on both desktop and laptop NVIDIA RTX 4090 GPUs. Notably, on the laptop 4090, it achieves a total 10.1× speedup by eliminating CPU offloading. Our NVFP4 model is also 3.1× faster than both BF16 and NF4 on the RTX 5090 GPU.
    }
    \lblfig{efficiency}
    \vspace{-15pt}
\end{figure}
\begin{figure}[t]
    \centering
    \vspace{-15pt}
    \includegraphics[width=\linewidth]{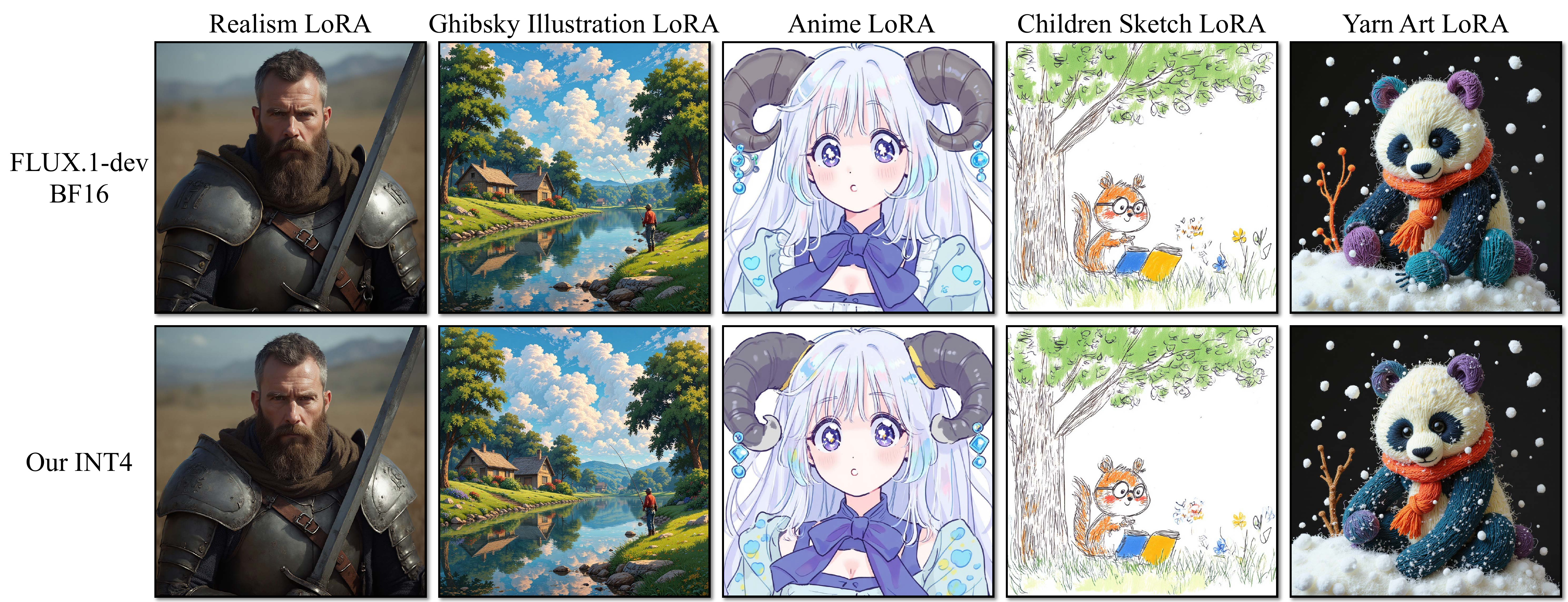}
    \vspace{-15pt}
	\caption{
		\looseness=-1
        Our 4-bit model seamlessly integrates with off-the-shelf LoRAs without requiring requantization. When applying LoRAs, it matches the image quality of the original 16-bit FLUX.1-dev. See \app{Text Prompts} for the text prompts.
    }
    \vspace{-10pt}
    \lblfig{lora}
\end{figure}
\begin{figure}[t]
    \centering
    \includegraphics[width=\linewidth]{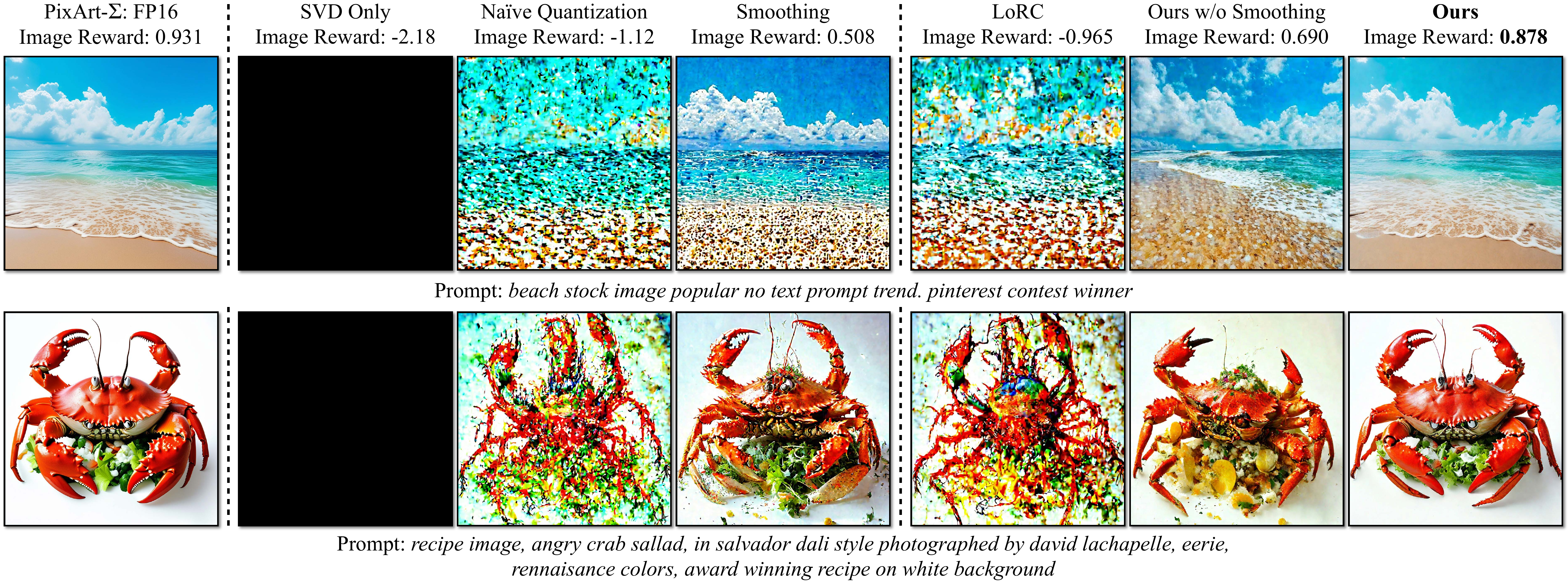}
    \vspace{-15pt}
	\caption{
		Ablation study of \method on PixArt-$\Sigma$. The rank of the low-rank branch is 64. Image Reward is measured over 1K samples from MJHQ. Our results significantly outperform the others, achieving the highest image quality by a wide margin.
    }
    \vspace{-5pt}
    \lblfig{ablation}
\end{figure}

\myparagraph{Memory save and speedup.}
In \fig{efficiency}, we report measured model size, memory savings, and speedup for FLUX.1. Our INT4 and NVFP4 quantization reduce the original transformer size from 22.2 GiB to 6.1 GiB, including a 0.3 GiB overhead due to the low-rank branch, resulting in an overall 3.6× reduction. Since both weights and activations are quantized, compared to the NF4 weight-only-quantized variant, our inference engine \engine even saves more memory footprint. It offers a 3.0× speedup on both desktop- and laptop-level NVIDIA RTX 4090 GPUs with INT4 precision and a 3.1× speedup on the RTX 5090 GPU with NVFP4 precision, compared to both NF4 and the original 16-bit models. Notably, while the original BF16 model requires per-layer CPU offloading on the 16GB laptop 4090, our INT4 model fits entirely in GPU memory, resulting in a 10.1× speedup by avoiding offloading.

\looseness=-1
\myparagraph{Integrate with LoRA.} Previous quantization methods require fusing the LoRA branches and re-quantizing the model when integrating LoRAs. In contrast, our \engine eliminates redundant memory access, allowing adding a separate LoRA branch. In practice, we can fuse the LoRA branch into our low-rank branch by slightly increasing the rank, further enhancing efficiency. In \fig{lora}, we exhibit some visual examples of applying LoRAs of five different styles (\href{https://huggingface.co/XLabs-AI/flux-RealismLora}{Realism}, \href{https://huggingface.co/aleksa-codes/flux-ghibsky-illustration}{Ghibsky Illustration}, \href{https://huggingface.co/alvdansen/sonny-anime-fixed}{Anime}, \href{https://huggingface.co/Shakker-Labs/FLUX.1-dev-LoRA-Children-Simple-Sketch}{Children Sketch}, and \href{https://huggingface.co/linoyts/yarn_art_Flux_LoRA}{Yarn Art}) to our INT4 FLUX.1-dev model. Our INT4 model successfully adapts to each style while preserving the image quality of the 16-bit version. For more visual examples, see \app{LoRA Results}. For FLUX.1-schnell, we further support LoRAs from one-step conditional model pix2pix-turbo~\citep{parmar2024one}, enabling additional controls like sketch. An interactive demo is available \href{https://svdquant.mit.edu/flux1-schnell-sketch/}{here}.

\looseness=-1
\myparagraph{Ablation study.}
In \fig{ablation}, we present several ablation studies of \method on PixArt-$\Sigma$. First, both SVD-only and \naive quantization perform poorly in the 4-bit setting, resulting in severe quality degradation. While applying smoothing to the quantization slightly improves image quality compared to \naive quantization, the results remain unsatisfactory. LoRC~\citep{yao2023zeroquant} introduces a low-rank branch to compensate for quantization errors, but this approach is suboptimal, as quantization errors exhibit a well-spread distribution of singular values. Consequently, low-rank compensation fails to effectively mitigate these errors, as discussed in \sect{SVDQuant}. In contrast, we first decompose the weights and quantize only the residual. As demonstrated in \fig{singular_values}, the first several singular values are significantly larger than the rest, allowing us to shift them to the low-rank branch, effectively reducing weight magnitude. Finally, smoothing consolidates the outliers, enabling the low-rank branch to absorb outliers from the activations and substantially improving image quality. \lblsect{Ablation study.}

\myparagraph{Trade-off of increasing rank.} Please refer to \app{Trade-off of Increasing Rank} for more details.

\section{Conclusion}
\looseness=-1
In this work, we introduce a novel 4-bit post-training quantization paradigm, \method, for diffusion models. It adopts a low-rank branch to absorb the outliers in both the weights and activations, easing the process of quantization. Our inference engine \engine further fuses the low-rank and low-bit branch kernels, reducing memory usage and cutting off redundant data movement overhead. Extensive experiments demonstrate that \method preserves image quality. \engine further achieves a 3.5× reduction in memory usage over the original 16-bit model and 3.0× speedup over the W4A16 on an NVIDIA RTX 4090 and 5090 GPUs. This advancement enables the efficient deployment of large-scale diffusion models on edge devices, unlocking broader potential for interactive AI applications.

\section*{Acknowledgments}
We thank MIT-IBM Watson AI Lab, MIT and Amazon Science Hub, MIT AI Hardware Program, National Science Foundation, Packard Foundation, Dell, LG, Hyundai, and Samsung for supporting this research. We thank NVIDIA for donating the DGX server.

\subsubsection*{Changelog}
\noindent\textbf{V1} Initial preprint release.

\noindent\textbf{V2} Fix a typo.

\noindent\textbf{V3} ICLR 2025 camera-ready version. Upgrade the models by combining SVDQuant and GPTQ. Update NVFP4 and SANA results.

\bibliography{main}

@STRING{CVPR = {CVPR}}

@STRING{ICCV = {ICCV}}

@STRING{AAAI = {AAAI}}

@STRING{ECCV = {ECCV}}

@STRING{ICLR = {ICLR}}

@STRING{ICML = {ICML}}

@STRING{NeurIPS = {NeurIPS}}

@STRING{PMLR = {PMLR}}

@STRING{MLSys = {MLSys}}

@inproceedings{zhang2022gddim,
      title={gDDIM: Generalized denoising diffusion implicit models}, 
      author={Qinsheng Zhang and Molei Tao and Yongxin Chen},
      year={2022},
      booktitle=ICLR,
}

@inproceedings{song2023consistency,
  title={Consistency models},
  author={Song, Yang and Dhariwal, Prafulla and Chen, Mark and Sutskever, Ilya},
  booktitle=ICML,
  year={2023}
}

@inproceedings{salimans2021progressive,
  title={Progressive Distillation for Fast Sampling of Diffusion Models},
  author={Salimans, Tim and Ho, Jonathan},
  booktitle=ICLR,
  year={2021}
}

@inproceedings{he2016deep,
  title={Deep residual learning for image recognition},
  author={He, Kaiming and Zhang, Xiangyu and Ren, Shaoqing and Sun, Jian},
  booktitle=CVPR,
  year={2016}
}

@inproceedings{sohl2015deep,
  title={Deep unsupervised learning using nonequilibrium thermodynamics},
  author={Sohl-Dickstein, Jascha and Weiss, Eric and Maheswaranathan, Niru and Ganguli, Surya},
  booktitle=ICML,
  year={2015},
}

@article{ho2020denoising,
  title={Denoising diffusion probabilistic models},
  author={Ho, Jonathan and Jain, Ajay and Abbeel, Pieter},
  journal=NeurIPS,
  year={2020}
}

@inproceedings{
      meng2022sdedit,
      title={{SDE}dit: Guided Image Synthesis and Editing with Stochastic Differential Equations},
      author={Chenlin Meng and Yutong He and Yang Song and Jiaming Song and Jiajun Wu and Jun-Yan Zhu and Stefano Ermon},
      booktitle=ICLR,
      year={2022},
}

@inproceedings{li2020gan,
  title={Gan compression: Efficient architectures for interactive conditional gans},
  author={Li, Muyang and Lin, Ji and Ding, Yaoyao and Liu, Zhijian and Zhu, Jun-Yan and Han, Song},
  booktitle=CVPR,
  year={2020}
}

@inproceedings{li2022efficient,
  title={Efficient Spatially Sparse Inference for Conditional GANs and Diffusion Models},
  author={Li, Muyang and Lin, Ji and Meng, Chenlin and Ermon, Stefano and Han, Song and Zhu, Jun-Yan},
  booktitle=NeurIPS,
  year={2022}
}

@inproceedings{zhang2018perceptual,
  title={The Unreasonable Effectiveness of Deep Features as a Perceptual Metric},
  author={Zhang, Richard and Isola, Phillip and Efros, Alexei A and Shechtman, Eli and Wang, Oliver},
  booktitle=CVPR,
  year={2018}
}

@article{heusel2017gans,
  title={Gans trained by a two time-scale update rule converge to a local nash equilibrium},
  author={Heusel, Martin and Ramsauer, Hubert and Unterthiner, Thomas and Nessler, Bernhard and Hochreiter, Sepp},
  journal=NeurIPS,
  year={2017}
}

@inproceedings{parmar2021cleanfid,
  title={On Aliased Resizing and Surprising Subtleties in GAN Evaluation},
  author={Parmar, Gaurav and Zhang, Richard and Zhu, Jun-Yan},
  booktitle=CVPR,
  year={2022}
}

@inproceedings{rombach2022high,
  title={High-resolution image synthesis with latent diffusion models},
  author={Rombach, Robin and Blattmann, Andreas and Lorenz, Dominik and Esser, Patrick and Ommer, Bj{\"o}rn},
  booktitle=CVPR,
  year={2022}
}

@inproceedings{lu2022dpm,
  title={Dpm-solver: A fast ode solver for diffusion probabilistic model sampling in around 10 steps},
  author={Lu, Cheng and Zhou, Yuhao and Bao, Fan and Chen, Jianfei and Li, Chongxuan and Zhu, Jun},
  booktitle=NeurIPS,
  year={2022}
}

@article{meng2022distillation,
  title={On distillation of guided diffusion models},
  author={Meng, Chenlin and Gao, Ruiqi and Kingma, Diederik P and Ermon, Stefano and Ho, Jonathan and Salimans, Tim},
  journal={arXiv preprint arXiv:2210.03142},
  year={2022}
}

@inproceedings{ronneberger2015u,
  title={U-net: Convolutional networks for biomedical image segmentation},
  author={Ronneberger, Olaf and Fischer, Philipp and Brox, Thomas},
  booktitle={Medical Image Computing and Computer-Assisted Intervention--MICCAI 2015: 18th International Conference, Munich, Germany, October 5-9, 2015, Proceedings, Part III 18},
  pages={234--241},
  year={2015},
  organization={Springer}
}

@inproceedings{li2023q,
  title={Q-diffusion: Quantizing diffusion models},
  author={Li, Xiuyu and Liu, Yijiang and Lian, Long and Yang, Huanrui and Dong, Zhen and Kang, Daniel and Zhang, Shanghang and Keutzer, Kurt},
  booktitle=ICCV,
  year={2023}
}

@inproceedings{podell2023sdxl,
  title={SDXL: Improving Latent Diffusion Models for High-Resolution Image Synthesis},
  author={Podell, Dustin and English, Zion and Lacey, Kyle and Blattmann, Andreas and Dockhorn, Tim and M{\"u}ller, Jonas and Penna, Joe and Rombach, Robin},
  booktitle={ICLR},
  year={2024}
}

@article{balaji2022ediffi,
  title={ediffi: Text-to-image diffusion models with an ensemble of expert denoisers},
  author={Balaji, Yogesh and Nah, Seungjun and Huang, Xun and Vahdat, Arash and Song, Jiaming and Kreis, Karsten and Aittala, Miika and Aila, Timo and Laine, Samuli and Catanzaro, Bryan and others},
  journal={arXiv preprint arXiv:2211.01324},
  year={2022}
}

@inproceedings{zhang2022fast,
  title={Fast Sampling of Diffusion Models with Exponential Integrator},
  author={Zhang, Qinsheng and Chen, Yongxin},
  booktitle=ICLR,
  year={2022}
}

@article{li2023snapfusion,
  title={SnapFusion: Text-to-Image Diffusion Model on Mobile Devices within Two Seconds},
  author={Li, Yanyu and Wang, Huan and Jin, Qing and Hu, Ju and Chemerys, Pavlo and Fu, Yun and Wang, Yanzhi and Tulyakov, Sergey and Ren, Jian},
  journal=NeurIPS,
  year={2023}
}

@article{chen2015microsoft,
  title={Microsoft coco captions: Data collection and evaluation server},
  author={Chen, Xinlei and Fang, Hao and Lin, Tsung-Yi and Vedantam, Ramakrishna and Gupta, Saurabh and Doll{\'a}r, Piotr and Zitnick, C Lawrence},
  journal={arXiv preprint arXiv:1504.00325},
  year={2015}
}

@article{luo2023latent,
  title   = {Latent Consistency Models: Synthesizing High-Resolution Images with Few-Step Inference},
  author  = {Simian Luo and Yiqin Tan and Longbo Huang and Jian Li and Hang Zhao},
  year    = {2023},
  journal = {arXiv preprint arXiv: 2310.04378}
}

@inproceedings{xiao2023smoothquant,
  title={Smoothquant: Accurate and efficient post-training quantization for large language models},
  author={Xiao, Guangxuan and Lin, Ji and Seznec, Mickael and Wu, Hao and Demouth, Julien and Han, Song},
  booktitle=ICML,
  year={2023},
}

@inproceedings{lin2024awq,
  title={AWQ: Activation-aware Weight Quantization for On-Device LLM Compression and Acceleration},
  author={Lin, Ji and Tang, Jiaming and Tang, Haotian and Yang, Shang and Chen, Wei-Ming and Wang, Wei-Chen and Xiao, Guangxuan and Dang, Xingyu and Gan, Chuang and Han, Song},
  booktitle=MLSys,
  year={2024}
}

@inproceedings{lin2025qserve,
  title={Qserve: W4a8kv4 quantization and system co-design for efficient llm serving},
  author={Lin, Yujun and Tang, Haotian and Yang, Shang and Zhang, Zhekai and Xiao, Guangxuan and Gan, Chuang and Han, Song},
  booktitle=MLSys,
  year={2025}
}

@article{dettmers2022gpt3,
  title={Gpt3. int8 (): 8-bit matrix multiplication for transformers at scale},
  author={Dettmers, Tim and Lewis, Mike and Belkada, Younes and Zettlemoyer, Luke},
  journal=NeurIPS,
  year={2022}
}

@inproceedings{zhao2024mixdq,
  title={Mixdq: Memory-efficient few-step text-to-image diffusion models with metric-decoupled mixed precision quantization},
  author={Zhao, Tianchen and Ning, Xuefei and Fang, Tongcheng and Liu, Enshu and Huang, Guyue and Lin, Zinan and Yan, Shengen and Dai, Guohao and Wang, Yu},
  booktitle=ECCV,
  year={2024},
}

@article{zhao2024vidit,
  title={ViDiT-Q: Efficient and Accurate Quantization of Diffusion Transformers for Image and Video Generation},
  author={Zhao, Tianchen and Fang, Tongcheng and Liu, Enshu and Rui, Wan and Soedarmadji, Widyadewi and Li, Shiyao and Lin, Zinan and Dai, Guohao and Yan, Shengen and Yang, Huazhong and others},
  journal={arXiv preprint arXiv:2406.02540},
  year={2024}
}

@article{frantar-gptq,
  title={{GPTQ}: Accurate Post-training Compression for Generative Pretrained Transformers}, 
  author={Elias Frantar and Saleh Ashkboos and Torsten Hoefler and Dan Alistarh},
  year={2023},
  journal=ICLR
}

@inproceedings{sauer2023adversarial,
  title={Adversarial diffusion distillation},
  author={Sauer, Axel and Lorenz, Dominik and Blattmann, Andreas and Rombach, Robin},
  booktitle=ECCV,
  year={2024}
}

@inproceedings{peebles2023scalable,
  title={Scalable diffusion models with transformers},
  author={Peebles, William and Xie, Saining},
  booktitle=ICCV,
  year={2023}
}

@inproceedings{yin2024one,
  title={One-step diffusion with distribution matching distillation},
  author={Yin, Tianwei and Gharbi, Micha{\"e}l and Zhang, Richard and Shechtman, Eli and Durand, Fredo and Freeman, William T and Park, Taesung},
  booktitle=CVPR,
  year={2024}
}

@inproceedings{yin2024improved,
  title={Improved Distribution Matching Distillation for Fast Image Synthesis},
  author={Yin, Tianwei and Gharbi, Micha{\"e}l and Park, Taesung and Zhang, Richard and Shechtman, Eli and Durand, Fredo and Freeman, William T},
  booktitle=NeurIPS,
  year={2024}
}

@misc{li2024playground,
      title={Playground v2.5: Three Insights towards Enhancing Aesthetic Quality in Text-to-Image Generation}, 
      author={Daiqing Li and Aleks Kamko and Ehsan Akhgari and Ali Sabet and Linmiao Xu and Suhail Doshi},
      year={2024},
      eprint={2402.17245},
      archivePrefix={arXiv},
      primaryClass={cs.CV}
}

@inproceedings{urbanek2024picture,
  title={A picture is worth more than 77 text tokens: Evaluating clip-style models on dense captions},
  author={Urbanek, Jack and Bordes, Florian and Astolfi, Pietro and Williamson, Mary and Sharma, Vasu and Romero-Soriano, Adriana},
  booktitle=CVPR,
  year={2024}
}

@article{eckart1936approximation,
  title={The approximation of one matrix by another of lower rank},
  author={Eckart, Carl and Young, Gale},
  journal={Psychometrika},
  volume={1},
  number={3},
  pages={211--218},
  year={1936},
  publisher={Springer-Verlag}
}

@article{mirsky1960symmetric,
  title={Symmetric gauge functions and unitarily invariant norms},
  author={Mirsky, Leon},
  journal={The quarterly journal of mathematics},
  volume={11},
  number={1},
  pages={50--59},
  year={1960},
  publisher={Oxford University Press}
}

@inproceedings{li2024distrifusion,
  title={Distrifusion: Distributed parallel inference for high-resolution diffusion models},
  author={Li, Muyang and Cai, Tianle and Cao, Jiaxin and Zhang, Qinsheng and Cai, Han and Bai, Junjie and Jia, Yangqing and Liu, Ming-Yu and Li, Kai and Han, Song},
  booktitle=CVPR,
  year={2024}
}

@inproceedings{wang2022exploring,
    author = {Wang, Jianyi and Chan, Kelvin CK and Loy, Chen Change},
    title = {Exploring CLIP for Assessing the Look and Feel of Images},
    booktitle = AAAI,
    year = {2023}
}

@inproceedings{radford2021learning,
  title={Learning transferable visual models from natural language supervision},
  author={Radford, Alec and Kim, Jong Wook and Hallacy, Chris and Ramesh, Aditya and Goh, Gabriel and Agarwal, Sandhini and Sastry, Girish and Askell, Amanda and Mishkin, Pamela and Clark, Jack and others},
  booktitle=ICML,
  year={2021},
}

@proceedings{kang2024distilling,
  title={Distilling Diffusion Models into Conditional GANs},
  author={Kang, Minguk and Zhang, Richard and Barnes, Connelly and Paris, Sylvain and Kwak, Suha and Park, Jaesik and Shechtman, Eli and Zhu, Jun-Yan and Park, Taesung},
  booktitle=ECCV,
  year={2024}
}

@inproceedings{ma2024deepcache,
  title={Deepcache: Accelerating diffusion models for free},
  author={Ma, Xinyin and Fang, Gongfan and Wang, Xinchao},
  booktitle=CVPR,
  year={2024}
}

@inproceedings{cai2024condition,
  title={Condition-Aware Neural Network for Controlled Image Generation},
  author={Cai, Han and Li, Muyang and Zhang, Qinsheng and Liu, Ming-Yu and Han, Song},
  booktitle=CVPR,
  year={2024}
}

@article{zhao2024atom,
  title={Atom: Low-bit quantization for efficient and accurate llm serving},
  author={Zhao, Yilong and Lin, Chien-Yu and Zhu, Kan and Ye, Zihao and Chen, Lequn and Zheng, Size and Ceze, Luis and Krishnamurthy, Arvind and Chen, Tianqi and Kasikci, Baris},
  journal=MLSys,
  year={2024}
}

@inproceedings{shang2023post,
  title={Post-training quantization on diffusion models},
  author={Shang, Yuzhang and Yuan, Zhihang and Xie, Bin and Wu, Bingzhe and Yan, Yan},
  booktitle=CVPR,
  year={2023}
}

@article{ma2024learning,
  title={Learning-to-cache: Accelerating diffusion transformer via layer caching},
  author={Ma, Xinyin and Fang, Gongfan and Bi Mi, Michael and Wang, Xinchao},
  journal=NeurIPS,
  year={2024}
}

@article{he2023ptqd,
  title={Ptqd: Accurate post-training quantization for diffusion models},
  author={He, Yefei and Liu, Luping and Liu, Jing and Wu, Weijia and Zhou, Hong and Zhuang, Bohan},
  journal=NeurIPS,
  year={2023}
}

@inproceedings{huang2024tfmq,
  title={Tfmq-dm: Temporal feature maintenance quantization for diffusion models},
  author={Huang, Yushi and Gong, Ruihao and Liu, Jing and Chen, Tianlong and Liu, Xianglong},
  booktitle=CVPR,
  year={2024}
}

@inproceedings{heefficientdm,
  title={EfficientDM: Efficient Quantization-Aware Fine-Tuning of Low-Bit Diffusion Models},
  author={He, Yefei and Liu, Jing and Wu, Weijia and Zhou, Hong and Zhuang, Bohan},
  booktitle=ICLR,
  year={2024}
}

@article{yang2023efficient,
  title={Efficient quantization strategies for latent diffusion models},
  author={Yang, Yuewei and Dai, Xiaoliang and Wang, Jialiang and Zhang, Peizhao and Zhang, Hongbo},
  journal={arXiv preprint arXiv:2312.05431},
  year={2023}
}

@article{zheng2024binarydm,
  title={Binarydm: Towards accurate binarization of diffusion model},
  author={Zheng, Xingyu and Qin, Haotong and Ma, Xudong and Zhang, Mingyuan and Hao, Haojie and Wang, Jiakai and Zhao, Zixiang and Guo, Jinyang and Liu, Xianglong},
  journal={arXiv preprint arXiv:2404.05662},
  year={2024}
}

@article{liu2024enhanced,
  title={Enhanced distribution alignment for post-training quantization of diffusion models},
  author={Liu, Xuewen and Li, Zhikai and Xiao, Junrui and Gu, Qingyi},
  journal={arXiv preprint arXiv:2401.04585},
  year={2024}
}

@inproceedings{tang2023post,
  title={Post-training quantization with progressive calibration and activation relaxing for text-to-image diffusion models},
  author={Tang, Siao and Wang, Xin and Chen, Hong and Guan, Chaoyu and Wu, Zewen and Tang, Yansong and Zhu, Wenwu},
  booktitle=ECCV,
  year={2024},
}

@article{wang2024quest,
  title={Quest: Low-bit diffusion model quantization via efficient selective finetuning},
  author={Wang, Haoxuan and Shang, Yuzhang and Yuan, Zhihang and Wu, Junyi and Yan, Yan},
  journal={arXiv preprint arXiv:2402.03666},
  year={2024}
}

@inproceedings{sui2024bitsfusion,
  title={BitsFusion: 1.99 bits Weight Quantization of Diffusion Model},
  author={Sui, Yang and Li, Yanyu and Kag, Anil and Idelbayev, Yerlan and Cao, Junli and Hu, Ju and Sagar, Dhritiman and Yuan, Bo and Tulyakov, Sergey and Ren, Jian},
  booktitle=NeurIPS,
  year={2024}
}

@inproceedings{wang2024towards,
  title={Towards Accurate Post-training Quantization for Diffusion Models},
  author={Wang, Changyuan and Wang, Ziwei and Xu, Xiuwei and Tang, Yansong and Zhou, Jie and Lu, Jiwen},
  booktitle=CVPR,
  year={2024}
}

@inproceedings{wu2024ptq4dit,
  title={PTQ4DiT: Post-training Quantization for Diffusion Transformers},
  author={Wu, Junyi and Wang, Haoxuan and Shang, Yuzhang and Shah, Mubarak and Yan, Yan},
  booktitle=NeurIPS,
  year={2024}
}

@article{liu2024hq,
  title={HQ-DiT: Efficient Diffusion Transformer with FP4 Hybrid Quantization},
  author={Liu, Wenxuan and Zhang, Saiqian},
  journal={arXiv preprint arXiv:2405.19751},
  year={2024}
}

@article{ashkboos2024quarot,
  title={Quarot: Outlier-free 4-bit inference in rotated llms},
  author={Ashkboos, Saleh and Mohtashami, Amirkeivan and Croci, Maximilian and Li, Bo and Cameron, Pashmina and Jaggi, Martin and Alistarh, Dan and Hoefler, Torsten and Hensman, James},
  journal=NeurIPS,
  year={2024}
}

@article{liu2024spinquant,
  title={SpinQuant--LLM quantization with learned rotations},
  author={Liu, Zechun and Zhao, Changsheng and Fedorov, Igor and Soran, Bilge and Choudhary, Dhruv and Krishnamoorthi, Raghuraman and Chandra, Vikas and Tian, Yuandong and Blankevoort, Tijmen},
  journal={arXiv preprint arXiv:2405.16406},
  year={2024}
}

@InProceedings{kim2024squeezellm,
  title = 	 {{S}queeze{LLM}: Dense-and-Sparse Quantization},
  author =       {Kim, Sehoon and Hooper, Coleman Richard Charles and Gholami, Amir and Dong, Zhen and Li, Xiuyu and Shen, Sheng and Mahoney, Michael W. and Keutzer, Kurt},
  booktitle =ICML,
  year = 	 {2024},
}

@inproceedings{chen2024pixart,
  title={Pixart-$\sigma$: Weak-to-strong training of diffusion transformer for 4k text-to-image generation},
  author={Chen, Junsong and Ge, Chongjian and Xie, Enze and Wu, Yue and Yao, Lewei and Ren, Xiaozhe and Wang, Zhongdao and Luo, Ping and Lu, Huchuan and Li, Zhenguo},
  booktitle=ECCV,
  year={2024},
}

@inproceedings{esser2024scaling,
  title={Scaling rectified flow transformers for high-resolution image synthesis},
  author={Esser, Patrick and Kulal, Sumith and Blattmann, Andreas and Entezari, Rahim and M{\"u}ller, Jonas and Saini, Harry and Levi, Yam and Lorenz, Dominik and Sauer, Axel and Boesel, Frederic and others},
  booktitle=ICML,
  year={2024}
}

@article{chen2024asyncdiff,
  title={AsyncDiff: Parallelizing Diffusion Models by Asynchronous Denoising},
  author={Chen, Zigeng and Ma, Xinyin and Fang, Gongfan and Tan, Zhenxiong and Wang, Xinchao},
  journal={arXiv preprint arXiv:2406.06911},
  year={2024}
}

@article{wang2024pipefusion,
  title={PipeFusion: Displaced Patch Pipeline Parallelism for Inference of Diffusion Transformer Models},
  author={Wang, Jiannan and Fang, Jiarui and Li, Aoyu and Yang, PengCheng},
  journal={arXiv preprint arXiv:2405.14430},
  year={2024}
}

@inproceedings{dehghani2023scaling,
  title={Scaling vision transformers to 22 billion parameters},
  author={Dehghani, Mostafa and Djolonga, Josip and Mustafa, Basil and Padlewski, Piotr and Heek, Jonathan and Gilmer, Justin and Steiner, Andreas Peter and Caron, Mathilde and Geirhos, Robert and Alabdulmohsin, Ibrahim and others},
  booktitle=ICML,
  year={2023},
  organization={PMLR}
}

@inproceedings{bao2023all,
  title={All are worth words: A vit backbone for diffusion models},
  author={Bao, Fan and Nie, Shen and Xue, Kaiwen and Cao, Yue and Li, Chongxuan and Su, Hang and Zhu, Jun},
  booktitle=CVPR,
  year={2023}
}

@inproceedings{yao2023zeroquant,
  title={Exploring post-training quantization in LLMs from comprehensive study to low rank compensation},
  author={Yao, Zhewei and Wu, Xiaoxia and Li, Cheng and Youn, Stephen and He, Yuxiong},
  booktitle=AAAI,
  year={2024}
}

@article{guo2024lq,
  title={LQ-LoRA: Low-rank plus Quantized Matrix Decomposition for Efficient Language Model Finetuning},
  author={Guo, Han and Greengard, Philip and Xing, Eric and Kim, Yoon},
  journal=ICLR,
  year={2024},
}

@book{massart2007concentration,
  title={Concentration inequalities and model selection: Ecole d'Et{\'e} de Probabilit{\'e}s de Saint-Flour XXXIII-2003},
  author={Massart, Pascal},
  year={2007},
  publisher={Springer}
}

@inproceedings{lora,
    author = {Edward J. Hu and
    Yelong Shen and
    Phillip Wallis and
    Zeyuan Allen{-}Zhu and
    Yuanzhi Li and
    Shean Wang and
    Lu Wang and
    Weizhu Chen},
    booktitle = ICLR,
    title = {LoRA: Low-Rank Adaptation of Large Language Models},
    year = {2022}
}

@inproceedings{
    dettmers2023qlora,
    title={{QL}o{RA}: Efficient Finetuning of Quantized {LLM}s},
    author={Tim Dettmers and Artidoro Pagnoni and Ari Holtzman and Luke Zettlemoyer},
    booktitle=NeurIPS,
    year={2023},
}

@inproceedings{li2024loftq,
  author    = {Yixiao Li and Yifan Yu and Chen Liang and Nikos Karampatziakis and Pengcheng He and Weizhu Chen and Tuo Zhao},
  title     = {LoftQ: LoRA-Fine-Tuning-aware Quantization for Large Language Models},
  booktitle = ICLR,
  year      = {2024},
}

@inproceedings{xu2024qalora,
  author    = {Yuhui Xu and Lingxi Xie and Xiaotao Gu and Xin Chen and Heng Chang and Hengheng Zhang and Zhengsu Chen and Xiaopeng Zhang and Qi Tian},
  title     = {QA-LoRA: Quantization-Aware Low-Rank Adaptation of Large Language Models},
  booktitle = ICLR,
  year      = {2024},
}

@inproceedings{ruiz2023dreambooth,
  title={Dreambooth: Fine tuning text-to-image diffusion models for subject-driven generation},
  author={Ruiz, Nataniel and Li, Yuanzhen and Jampani, Varun and Pritch, Yael and Rubinstein, Michael and Aberman, Kfir},
  booktitle=CVPR,
  year={2023}
}

@inproceedings{zhang2023adding,
  title={Adding conditional control to text-to-image diffusion models},
  author={Zhang, Lvmin and Rao, Anyi and Agrawala, Maneesh},
  booktitle=ICCV,
  year={2023}
}

@misc{flux1,
  author = "Black-Forest-Labs",
  title = "FLUX.1",
  year = "2024",
  url = "https://blackforestlabs.ai/",
}

@misc{auraflow0.1,
  author = "fal.ai",
  title = "AuraFlow v0.1",
  year = "2024",
  url = "https://blog.fal.ai/auraflow/",
}

@article{liu2024linfusion,
  title={LinFusion: 1 GPU, 1 Minute, 16K Image},
  author={Liu, Songhua and Yu, Weihao and Tan, Zhenxiong and Wang, Xinchao},
  journal={arXiv preprint arXiv:2409.02097},
  year={2024}
}

@misc{
    Lllyasviel, 
    title={[major update] bitsandbytes guidelines and flux · Lllyasviel stable-diffusion-webui-forge · discussion \#981}, 
    url={https://github.com/lllyasviel/stable-diffusion-webui-forge/discussions/981}, 
    journal={GitHub}, 
    author={Lllyasviel},
    year={2024}
}

@inproceedings{hsu2022lowrank,
  author    = {Yen{-}Chang Hsu and Ting Hua and Sungen Chang and Qian Lou and Yilin Shen and Hongxia Jin},
  title     = {Language model compression with weighted low-rank factorization},
  booktitle = ICLR,
  year      = {2022},
}

@article{yuan2023asvd,
  title   = {ASVD: Activation-aware Singular Value Decomposition for Compressing Large Language Models},
  author  = {Zhihang Yuan and Yuzhang Shang and Yue Song and Qiang Wu and Yan Yan and Guangyu Sun},
  year    = {2023},
  journal = {arXiv preprint arXiv: 2312.05821}
}

@InProceedings{li2023losparse,
  title = 	 {{L}o{S}parse: Structured Compression of Large Language Models based on Low-Rank and Sparse Approximation},
  author =       {Li, Yixiao and Yu, Yifan and Zhang, Qingru and Liang, Chen and He, Pengcheng and Chen, Weizhu and Zhao, Tuo},
  booktitle = 	 ICML,
  year = 	 {2023},
  volume = 	 {202},
  publisher =    {PMLR},
}

@InProceedings{zhao2024galore,
  title = 	 {{G}a{L}ore: Memory-Efficient {LLM} Training by Gradient Low-Rank Projection},
  author =       {Zhao, Jiawei and Zhang, Zhenyu and Chen, Beidi and Wang, Zhangyang and Anandkumar, Anima and Tian, Yuandong},
  booktitle = 	 ICML,
  year = 	 {2024}
}

@article{jaiswal2024welore,
  title   = {From GaLore to WeLore: How Low-Rank Weights Non-uniformly Emerge from Low-Rank Gradients},
  author  = {Ajay Jaiswal and Lu Yin and Zhenyu Zhang and Shiwei Liu and Jiawei Zhao and Yuandong Tian and Zhangyang Wang},
  year    = {2024},
  journal = {arXiv preprint arXiv: 2407.11239}
}

@article{xu2024imagereward,
  title={Imagereward: Learning and evaluating human preferences for text-to-image generation},
  author={Xu, Jiazheng and Liu, Xiao and Wu, Yuchen and Tong, Yuxuan and Li, Qinkai and Ding, Ming and Tang, Jie and Dong, Yuxiao},
  journal=NeurIPS,
  year={2024}
}

@article{hessel2021clipscore,
  title={Clipscore: A reference-free evaluation metric for image captioning},
  author={Hessel, Jack and Holtzman, Ari and Forbes, Maxwell and Bras, Ronan Le and Choi, Yejin},
  journal={arXiv preprint arXiv:2104.08718},
  year={2021}
}

@misc{blackwell,
    title={NVIDIA Blackwell Architecture Technical Brief},
    year= {2024},
    author = {NVIDIA},
    url={https://resources.nvidia.com/en-us-blackwell-architecture}
}

@article{wang2023bitnet,
  title={Bitnet: Scaling 1-bit transformers for large language models},
  author={Wang, Hongyu and Ma, Shuming and Dong, Li and Huang, Shaohan and Wang, Huaijie and Ma, Lingxiao and Yang, Fan and Wang, Ruiping and Wu, Yi and Wei, Furu},
  journal={arXiv preprint arXiv:2310.11453},
  year={2023}
}

@article{ma2024era,
  title={The era of 1-bit llms: All large language models are in 1.58 bits},
  author={Ma, Shuming and Wang, Hongyu and Ma, Lingxiao and Wang, Lei and Wang, Wenhui and Huang, Shaohan and Dong, Li and Wang, Ruiping and Xue, Jilong and Wei, Furu},
  journal={arXiv preprint arXiv:2402.17764},
  year={2024}
}

@article{meng2024pissa,
  title={Pissa: Principal singular values and singular vectors adaptation of large language models},
  author={Meng, Fanxu and Wang, Zhaohui and Zhang, Muhan},
  journal=NeurIPS,
  year={2024}
}

@article{parmar2024one,
  title={One-step image translation with text-to-image models},
  author={Parmar, Gaurav and Park, Taesung and Narasimhan, Srinivasa and Zhu, Jun-Yan},
  journal={arXiv preprint arXiv:2403.12036},
  year={2024}
}

@article{xie2024sana,
  title={Sana: Efficient high-resolution image synthesis with linear diffusion transformers},
  author={Xie, Enze and Chen, Junsong and Chen, Junyu and Cai, Han and Tang, Haotian and Lin, Yujun and Zhang, Zhekai and Li, Muyang and Zhu, Ligeng and Lu, Yao and others},
  journal=ICLR,
  year={2025}
}

@article{chen2024deep,
  title={Deep compression autoencoder for efficient high-resolution diffusion models},
  author={Chen, Junyu and Cai, Han and Chen, Junsong and Xie, Enze and Yang, Shang and Tang, Haotian and Li, Muyang and Lu, Yao and Han, Song},
  journal=ICLR,
  year={2025}
}

@manual{nvidia2025blockscaling,
  title        = {Block Scaling in cuDNN Frontend API},
  author       = {{NVIDIA Corporation}},
  year         = {2025},
  url          = {https://docs.nvidia.com/deeplearning/cudnn/frontend/latest/operations/BlockScaling.html},
}
\bibliographystyle{iclr2025_conference}

\newpage
\appendix
\section{Proofs}
\setcounter{proposition}{0} 
\renewcommand{\theproposition}{4.\arabic{proposition}}
\subsection{Proof of \pps{decomp}}

\begin{proposition}
The quantization error $E(\mX, \mW) = \norm{\mX \mW - Q(\mX)Q(\mW)}_F$ in \eqn{error_def} can be decomposed as follows:
\begin{align}
    E(\mX, \mW) \le \norm{\mX}_F\norm{\mW - Q(\mW)}_F + \norm{\mX - Q(\mX)}_F(\norm{\mW}_F+\norm{\mW-Q(\mW)}_F).
\end{align}
\end{proposition}
\lblapp{proof1}
\begin{proof}
\begin{align*}
    & \norm{\mX \mW - Q(\mX)Q(\mW)}_F \\
  = & \norm{\mX \mW - \mX Q(\mW) + \mX Q(\mW) - Q(\mX)Q(\mW)}_F \\
\le & \norm{\mX(\mW - Q(\mW))}_F + \norm{(\mX - Q(\mX))Q(\mW)}_F \\
\le & \norm{\mX}_F\norm{\mW - Q(\mW)}_F + \norm{\mX - Q(\mX)}_F\norm{Q(\mW)}_F \\
\le & \norm{\mX}_F\norm{\mW - Q(\mW)}_F + \norm{\mX - Q(\mX)}_F\norm{\mW - (\mW - Q(\mW))}_F \\
\le & \norm{\mX}_F\norm{\mW - Q(\mW)}_F + \norm{\mX - Q(\mX)}_F(\norm{\mW}_F + \norm{\mW - Q(\mW)}_F).
\end{align*}
\end{proof}

\subsection{Proof of \pps{quant}}
\begin{proposition}
For any tensor $\mR$ and quantization method described in \eqn{quantization_def} as $Q(\mR) = s_\mR \cdot \mQ_\mR $. Assuming the elements of $\mR$ follow a distribution that satisfies the following regularity condition: There exists a constant $c$ such that
\begin{align}
    \mathbb{E} \sbr{\max(|\mR|)} \le c\cdot\mathbb{E} \sbr{\norm{\mR}_F}. \lbleqn{app-regularity}
\end{align}
Then, we have
\begin{align}
    \mathbb{E} \sbr{\norm{\mR - Q(\mR)}_F} \le \frac{c\sqrt{\text{size}(\mR)}}{q_{\max}} \cdot \mathbb{E}\sbr{\norm{\mR}_F}
\end{align}
where $\text{size}(\mR)$ denotes the number of elements in $\mR$. Especially if the elements of $\mR$ follow a normal distribution, \eqn{app-regularity} holds for $c = \sqrt{\frac{\log \rbr{\text{size}(\mR)}\pi}{\text{size}(\mR)}}$.
\end{proposition}
\lblapp{proof2}
\begin{proof}
\begin{align*}
    & \norm{\mR - Q(\mR) }_F \\
  = & \norm{\mR - s_\mR \cdot \mQ_\mR }_F \\
  = & \norm{s_\mR\cdot \frac{\mR}{s_{\mR}} -s_\mR \cdot \text{round}\left(\frac{\mR}{s_{\mR}}\right)}_F \\
  = & |s_\mR| \norm{\frac{\mR}{s_{\mR}} - \text{round}\left(\frac{\mR}{s_{\mR}}\right)}_F.
\end{align*}
So,
\begin{align}
    & \mathbb{E} \sbr{\norm{\mR - Q(\mR)}_F} \nonumber \\
\le & \mathbb{E} \sbr{|s_\mR|}\sqrt{\text{size}(\mR)} \nonumber \\
  = & \frac{\sqrt{\text{size}(\mR)}}{q_{\max}} \cdot \mathbb{E}\sbr{\max (|\mR|)} \nonumber \\
\le & \frac{c\sqrt{\text{size}(\mR)}}{q_{\max}} \cdot \mathbb{E}\sbr{\norm{\mR}_F} \nonumber
\end{align}

Especially, if the elements of $\mR$ follows a normal distribution, we have
\begin{align}
 \mathbb{E}\sbr{\max (|\mR|)} \le \sigma \sqrt{2\log \rbr{\text{size}(\mR)}}  \lbleqn{max_gaussian}
\end{align}
where $\sigma$ is the std deviation of the normal distribution. \eqn{max_gaussian} comes from the maximal inequality of Gaussian variables (Lemma 2.3 in \citet{massart2007concentration}).

On the other hand, 
\begin{align}
    & \mathbb{E}\sbr{\norm{\mR}_F} \nonumber\\
  = & \mathbb{E}\sbr{\sqrt{\sum_{x\in\mR}x^2}} \nonumber \\
\ge & \mathbb{E}\sbr{\frac{\sum_{x\in\mR} |x|}{\sqrt{\text{size}(\mR)}}} \lbleqn{cauchy} \\
  = & \sigma\sqrt{\frac{2\text{size}(\mR)}{\pi}} \lbleqn{half_normal},
\end{align}
where \eqn{cauchy} comes from Cauchy-Schwartz inequality and \eqn{half_normal} comes from the expectation of half-normal distribution.

Together, we have that for a normal distribution,
\begin{align*}
&\mathbb{E}\sbr{\max (|\mR|)}\\
\le& \sigma \sqrt{2\log \rbr{\text{size}(\mR)}}\\
\le& \sqrt{\frac{\log \rbr{\text{size}(\mR)}\pi}{\text{size}(\mR)}}\mathbb{E}\sbr{\norm{\mR}_F}.
\end{align*}
In other words, \eqn{app-regularity} holds for $c = \sqrt{\frac{\log \rbr{\text{size}(\mR)}\pi}{\text{size}(\mR)}}$.
\end{proof}

\section{Benchmark Models}
\lblapp{Benchmark Models}
We benchmark our methods using the following six text-to-image models:
\begin{itemize}[leftmargin=*]
    \vspace{-5pt}
    \item FLUX.1~\citep{flux1} is the SoTA open-sourced DiT-based diffusion model. It consists of 19 joint attention blocks~\citep{esser2024scaling} and 38 parallel attention blocks~\citep{dehghani2023scaling}, totaling 12B parameters. We evaluate both the 50-step guidance-distilled (FLUX.1-dev) and 4-step timestep-distilled (FLUX.1-schnell) variants.
    \item PixArt-$\Sigma$~\citep{chen2024pixart} is another DiT-based model. Instead of using joint attention, it stacks 28 attention blocks composed of self-attention, cross-attention, and feed-forward layers, amounting to 600M parameters. We evaluate it on the default 20-step setting.
    \item SANA~\citep{xie2024sana} is a 1.6B DiT model. It utilizes a 32× compression autoencoder~\citep{chen2024deep} and replaces Softmax attention with linear attention to accelerate image generation.
    \item \looseness=-1 Stable Diffusion XL (SDXL) is a widely-used UNet-based model with 2.6B parameters~\citep{podell2023sdxl}. It predicts noise with three resolution scales. The highest-resolution stage is processed entirely by ResBlocks~\citep{he2016deep}, while the other two stages jointly use ResBlocks and attention layers. Like PixArt-$\Sigma$, SDXL uses cross-attention layers for text conditioning. We evaluate it in the 30-step setting, along with its 4-step distilled variant, SDXL-Turbo~\citep{sauer2023adversarial}.
\end{itemize}

\section{Benchmark Datasets}
\lblapp{Benchmark Datasets}
To assess the generalization capability of our method, we adopt two distinct prompt sets with varying styles for benchmarking: 
\begin{itemize}[leftmargin=*]
    \vspace{-5pt}
    \item MJHQ-30K~\citep{li2024playground} consists of 30K samples from Midjourney with 10 common categories, 3K samples each. We randomly select 5K prompts from this dataset to evaluate model performance on artistic image generation.
    \item Densely Captioned Images (DCI)~\citep{urbanek2024picture} is a dataset containing $\sim$8K images with detailed human-annotated captions, averaging over 1,000 words. For our experiments, we use the summarized version (sDCI), where captions are condensed to 77 tokens using large language models (LLMs) to accommodate diffusion models. Similarly, we randomly sample 5K prompts for realistic image generation.
\end{itemize}

\section{Implementation Details}
\lblapp{Implementation Details}
For the 8-bit setting, we use per-token dynamic activation quantization and per-channel weight quantization with a low-rank branch of rank 16. For the 4-bit setting, we adopt per-group symmetric quantization for both activations and weights, along with a low-rank branch of rank 32. INT4 quantization uses a group size of 64 with 16-bit scales. We use NVFP4 for FP4 quantization, which has native hardware support of group size of 16 with FP8 scales on Blackwell GPUs \citep{nvidia2025blockscaling}. We use GPTQ~\citep{frantar-gptq} to quantize the residual weights. For FLUX.1 models, the inputs of linear layers in adaptive normalization are kept in 16 bits (\ie, W4A16). For other models, key and value projections in the cross-attention are retained at 16 bits since their latency only covers less than 5\% of total runtime.

The smoothing factor $\lambda \in \mathbb R^{m}$ is a per-channel vector whose $i$-th element is computed as $\lambda_i = \max(|\mX_{:,i}|)^\alpha / \max(|\mW_{i,:}|)^{1-\alpha}$ following SmoothQuant \citep{xiao2023smoothquant} Here, $\mX \in \mathbb R^{b\times m}$ and $\mW \in \mathbb R^{m\times n}$. It is decided offline by searching for the best migration strength $\alpha$ for each layer to minimize the layer output mean squared error (MSE) after SVD on the calibration dataset. 

\clearpage
\section{Additional Results}
\lblapp{Additional Results}
\subsection{Visual Quality Results}
\lblapp{Quality Results}

We report extra quantitative quality results with additional metrics in \tbl{quality-appendix}. Specifically, CLIP IQA~\citep{wang2022exploring} and CLIP Score~\citep{hessel2021clipscore} assesses the image quality and text-image alignment with CLIP~\citep{radford2021learning}, respectively. Structural Similarity Index Measure (SSIM) is used to measure the luminance, contrast, and structure similarity of images produced by our 4-bit model against the original 16-bit model. We also visualize more qualitative comparsions in \figs{flux-dev-appendix}, \ref{fig:flux-schnell-appendix}, \ref{fig:pixart-sigma-appendix}, \ref{fig:sdxl-appendix} and \ref{fig:sdxl-turbo-appendix}.

\renewcommand \arraystretch{1.}
\begin{table}[H]
    \setlength{\tabcolsep}{4pt}
    \caption{
        Additional quantitative quality comparisons across different models. RTN stands for round-to-nearest. C.IQA means CLIP IQA, and C.SCR means CLIP Score.
    }
    \vspace{-5pt}
    \lbltbl{quality-appendix}
    \scriptsize \centering
    \begin{tabular}{cccccccccc}
    \toprule
    & & & & \multicolumn{3}{c}{MJHQ} & \multicolumn{3}{c}{sDCI} \\
    \cmidrule(lr){5-7} \cmidrule(lr){8-10}
    Backbone & Model & Precision & Method & \multicolumn{2}{c}{Quality} & \multicolumn{1}{c}{Similarity} & \multicolumn{2}{c}{Quality} & \multicolumn{1}{c}{Similarity} \\
    \cmidrule(lr){5-6} \cmidrule(lr){7-7} \cmidrule(lr){8-9} \cmidrule(lr){10-10}
    & & & & C.IQA ($\uparrow$) & C.SCR ($\uparrow$) & SSIM( $\uparrow$) & C.IQA ($\uparrow$) & C.SCR ($\uparrow$) & SSIM ($\uparrow$) \\
    
    \midrule
    \multirow{30}{*}{DiT}& \multirow{7}{*}{\makecell{FLUX.1\\-dev\\(50 Steps)}} & BF16 & -- & 0.952 & 26.0 & -- & 0.955 & 25.4 & -- \\
    \cmidrule{3-10}
    & & \cellcolor{mitblue}INT W8A8 & \cellcolor{mitblue}Ours & \cellcolor{mitblue}0.953 & \cellcolor{mitblue}26.0 & \cellcolor{mitblue}0.748 & \cellcolor{mitblue}0.955 & \cellcolor{mitblue}25.4 & \cellcolor{mitblue}0.697   \\
    \cmidrule{3-10}
    & & W4A16 & NF4 & 0.947 & \textbf{25.8} & 0.748 & 0.951 & \textbf{25.4} & 0.697  \\
    & & \cellcolor{mitblue}INT W4A4 & \cellcolor{mitblue}Ours & \cellcolor{mitblue}0.950 & \cellcolor{mitblue}\textbf{25.8} & \cellcolor{mitblue}0.797 & \cellcolor{mitblue}0.951 & \cellcolor{mitblue}25.3 & \cellcolor{mitblue}0.751 \\
    & & \cellcolor{mitblue}NVFP W4A4 & \cellcolor{mitblue}Ours & \cellcolor{mitblue}\textbf{0.952} & \cellcolor{mitblue}\textbf{25.8} & \cellcolor{mitblue}\textbf{0.808} & \cellcolor{mitblue}\textbf{0.955} & \cellcolor{mitblue}\textbf{25.4} & \cellcolor{mitblue}\textbf{0.768} \\
    
    \cmidrule{2-10}
    & \multirow{7}{*}{\makecell{FLUX.1\\-schnell\\(4 Steps)}} & BF16  & -- & 0.938 & 26.6 & -- & 0.932 & 26.2 & -- \\
    \cmidrule{3-10}
    & & \cellcolor{mitblue}INT W8A8 & \cellcolor{mitblue}Ours & \cellcolor{mitblue}0.938 & \cellcolor{mitblue}26.6 & \cellcolor{mitblue}0.844 & \cellcolor{mitblue}0.932 & \cellcolor{mitblue}26.2 & \cellcolor{mitblue}0.811 \\
    \cmidrule{3-10}
    & & W4A16 & NF4 & \textbf{0.941} & \textbf{26.6} & 0.713 & \textbf{0.933} & \textbf{26.2} & 0.674 \\
    & & \cellcolor{mitblue}INT W4A4 & \cellcolor{mitblue}Ours & \cellcolor{mitblue}0.937 & \cellcolor{mitblue}26.5 & \cellcolor{mitblue}0.720 & \cellcolor{mitblue}0.932 & \cellcolor{mitblue}\textbf{26.2} & \cellcolor{mitblue}0.681 \\
    & & \cellcolor{mitblue}NVFP W4A4 & \cellcolor{mitblue}Ours & \cellcolor{mitblue}0.939 & \cellcolor{mitblue}\textbf{26.6} & \cellcolor{mitblue}\textbf{0.745} & \cellcolor{mitblue}0.932 & \cellcolor{mitblue}26.1 & \cellcolor{mitblue}0.712\\
    
    \cmidrule{2-10}
    & \multirow{8}{*}{\makecell{PixArt-$\Sigma$\\(20 Steps)}} & FP16 & -- & 0.944 & 26.8 & -- & 0.966 & 26.1 & -- \\
    \cmidrule{3-10}
    & & INT W8A8 & ViDiT-Q & \textbf{0.948} & 26.7 & 0.815 & 0.966 & \textbf{26.1} & 0.756 \\
    & & \cellcolor{mitblue}INT W8A8 & \cellcolor{mitblue}Ours & \cellcolor{mitblue}0.947 & \cellcolor{mitblue}\textbf{26.8} & \cellcolor{mitblue}\textbf{0.849} & \cellcolor{mitblue}\textbf{0.967} & \cellcolor{mitblue}26.0 & \cellcolor{mitblue}\textbf{0.800} \\
    \cmidrule{3-10}
    & & INT W4A8 & ViDiT-Q & 0.912 & 25.7 & 0.356 & 0.917 & 25.4 & 0.295 \\
    & & INT W4A4 & ViDiT-Q & 0.185 & 13.3 & 0.077 & 0.176 & 13.3 & 0.080 \\
    & & \cellcolor{mitblue}INT W4A4 & \cellcolor{mitblue}Ours & \cellcolor{mitblue}0.926 & \cellcolor{mitblue}26.6 & \cellcolor{mitblue}0.655 & \cellcolor{mitblue}0.948 & \cellcolor{mitblue}\textbf{26.1} & \cellcolor{mitblue}0.577  \\
    & & \cellcolor{mitblue}NVFP W4A4 & \cellcolor{mitblue}Ours & \cellcolor{mitblue}\textbf{0.938} & \cellcolor{mitblue}\textbf{26.7} & \cellcolor{mitblue}\textbf{0.692} & \cellcolor{mitblue}\textbf{0.956} & \cellcolor{mitblue}\textbf{26.1} & \cellcolor{mitblue}\textbf{0.618} \\    

    \cmidrule{2-10}
    & & BF16 & -- & 0.934 & 26.8 & -- & 0.958& 26.4 & --\\
    \cmidrule{3-10}
    & SANA& INT W4A4 & RTN & 0.915 & \textbf{26.9} & 0.604 & 0.943 & \textbf{26.4} & 0.538\\
    & -1.6B& \cellcolor{mitblue}INT W4A4 & \cellcolor{mitblue}Ours & \cellcolor{mitblue}0.926 & \cellcolor{mitblue}\textbf{26.9} & \cellcolor{mitblue}0.710 & \cellcolor{mitblue}0.951 & \cellcolor{mitblue}\textbf{26.4} & \cellcolor{mitblue}0.649  \\
    & (20 Steps)& NVFP W4A4 & RTN & 0.929 &  26.8 & 0.694 & 0.953 & \textbf{26.4 }& 0.626 \\
    & & \cellcolor{mitblue}NVFP W4A4 & \cellcolor{mitblue}Ours & \cellcolor{mitblue}\textbf{0.932} & \cellcolor{mitblue}\textbf{26.9} & \cellcolor{mitblue}\textbf{0.755} & \cellcolor{mitblue}\textbf{0.955} & \cellcolor{mitblue}\textbf{26.4} & \cellcolor{mitblue}\textbf{0.701} \\
    
    \midrule
    \multirow{15}{*}{UNet} & \multirow{9}{*}{\makecell{SDXL\\-Turbo\\(4 Steps)}} & FP16 & --& 0.926 & 26.5 & -- & 0.913 & 26.5 & -- \\
    \cmidrule{3-10}
    & & INT W8A8 & MixDQ & 0.922 & 26.5 & 0.763 & 0.907 & 26.5 & 0.750 \\
    & & \cellcolor{mitblue}INT W8A8 & \cellcolor{mitblue}Ours & \cellcolor{mitblue}\textbf{0.925} & \cellcolor{mitblue}\textbf{26.5} & \cellcolor{mitblue}\textbf{0.821} & \cellcolor{mitblue}\textbf{0.912} & \cellcolor{mitblue}\textbf{26.5} & \cellcolor{mitblue}\textbf{0.808} \\
    \cmidrule{3-10}
    & & INT W4A8 & MixDQ & 0.893 & 25.9 & 0.512 & \textbf{0.895} & 26.1 & 0.493 \\
    & & INT W4A4 & MixDQ & 0.556 & 13.1 & 0.289 & 0.548 & 11.9 & 0.296 \\
    & & \cellcolor{mitblue}INT W4A4 & \cellcolor{mitblue}Ours & \cellcolor{mitblue}0.915 & \cellcolor{mitblue}\textbf{26.5} & \cellcolor{mitblue}0.631 & \cellcolor{mitblue}0.894 & \cellcolor{mitblue}\textbf{26.8} & \cellcolor{mitblue}0.614 \\
    & & \cellcolor{mitblue}FP W4A4 & \cellcolor{mitblue}Ours & \cellcolor{mitblue}\textbf{0.919} & \cellcolor{mitblue}\textbf{26.5} & \cellcolor{mitblue}\textbf{0.663} & \cellcolor{mitblue}\textbf{0.902} & \cellcolor{mitblue}26.6 & \cellcolor{mitblue}\textbf{0.649} \\

    \cmidrule{2-10}
    & \multirow{6}{*}{\makecell{SDXL\\(30 Steps)}} & FP16 & -- & 0.907 & 27.2 & -- & 0.911 & 26.5 & -- \\
    \cmidrule{3-10}
    & & INT W8A8 & TensorRT & 0.905 & 26.7 & 0.733 & 0.901 & 26.1 & 0.697 \\
    & & \cellcolor{mitblue}INT W8A8 & \cellcolor{mitblue}Ours & \cellcolor{mitblue}\textbf{0.912} & \cellcolor{mitblue}\textbf{27.0} & \cellcolor{mitblue}\textbf{0.843} & \cellcolor{mitblue}\textbf{0.910} & \cellcolor{mitblue}\textbf{26.3} & \cellcolor{mitblue}\textbf{0.814} \\
    \cmidrule{3-10}
    & & \cellcolor{mitblue}INT W4A4 & \cellcolor{mitblue}Ours & \cellcolor{mitblue}0.878 & \cellcolor{mitblue}26.7 & \cellcolor{mitblue}0.717 & \cellcolor{mitblue}0.862 & \cellcolor{mitblue}26.2 & \cellcolor{mitblue}0.672 \\
    & & \cellcolor{mitblue}NVFP W4A4 & \cellcolor{mitblue}Ours & \cellcolor{mitblue}0.892 & \cellcolor{mitblue}\textbf{26.8} & \cellcolor{mitblue}\textbf{0.739} & \cellcolor{mitblue}0.877 & \cellcolor{mitblue}\textbf{26.4} & \cellcolor{mitblue}\textbf{0.701} \\ 
    \bottomrule
    \end{tabular}
\end{table}

\newpage
\begin{figure}[H]
    \centering
    \vspace{-20pt}
    \includegraphics[width=0.85\linewidth]{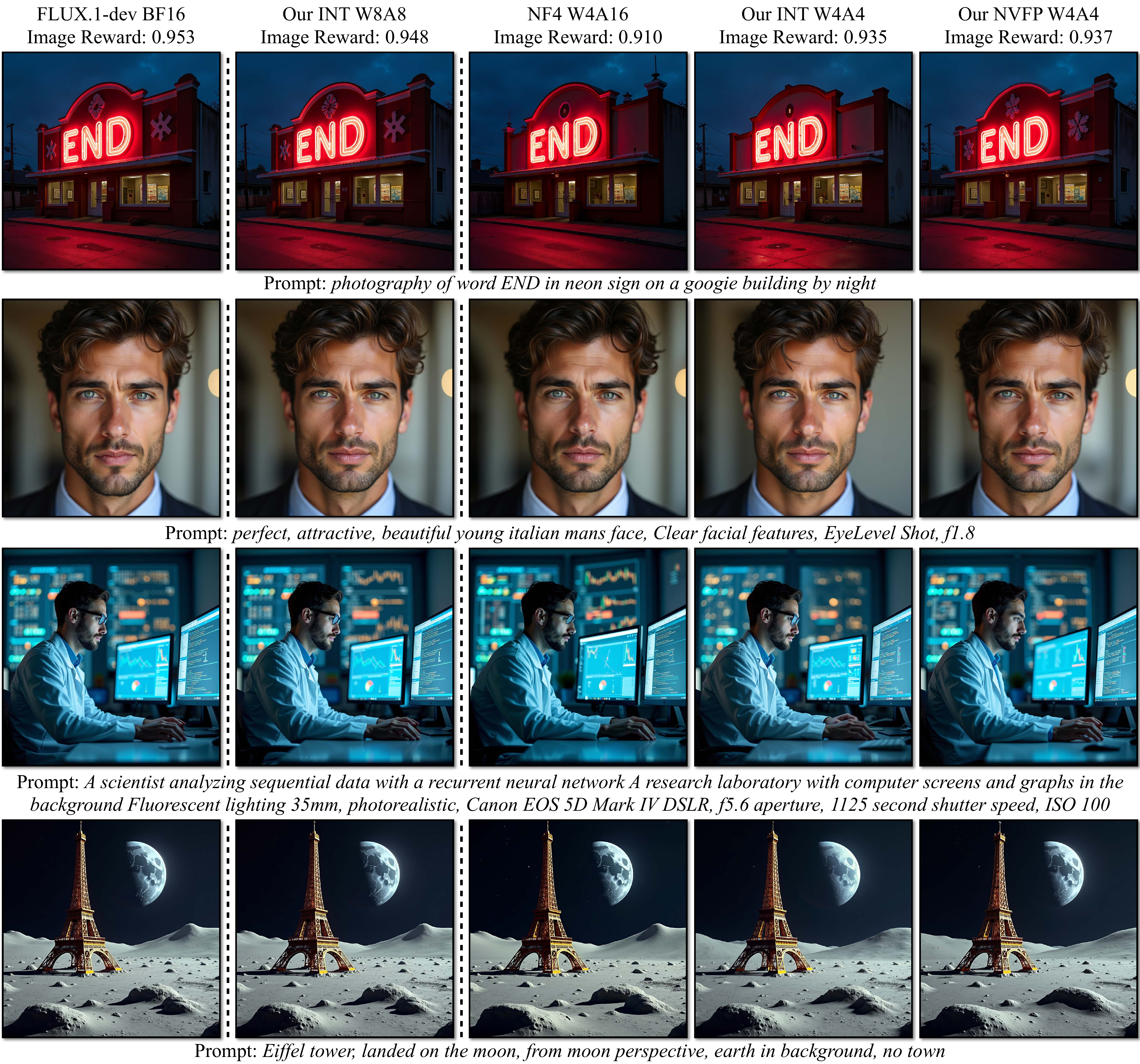}
    \vspace{-5pt}
	\caption{Qualitative visual results of FLUX.1-dev on MJHQ.}
    \vspace{-20pt}
    \lblfig{flux-dev-appendix}
\end{figure}
\begin{figure}[H]
    \centering
    \includegraphics[width=0.85\linewidth]{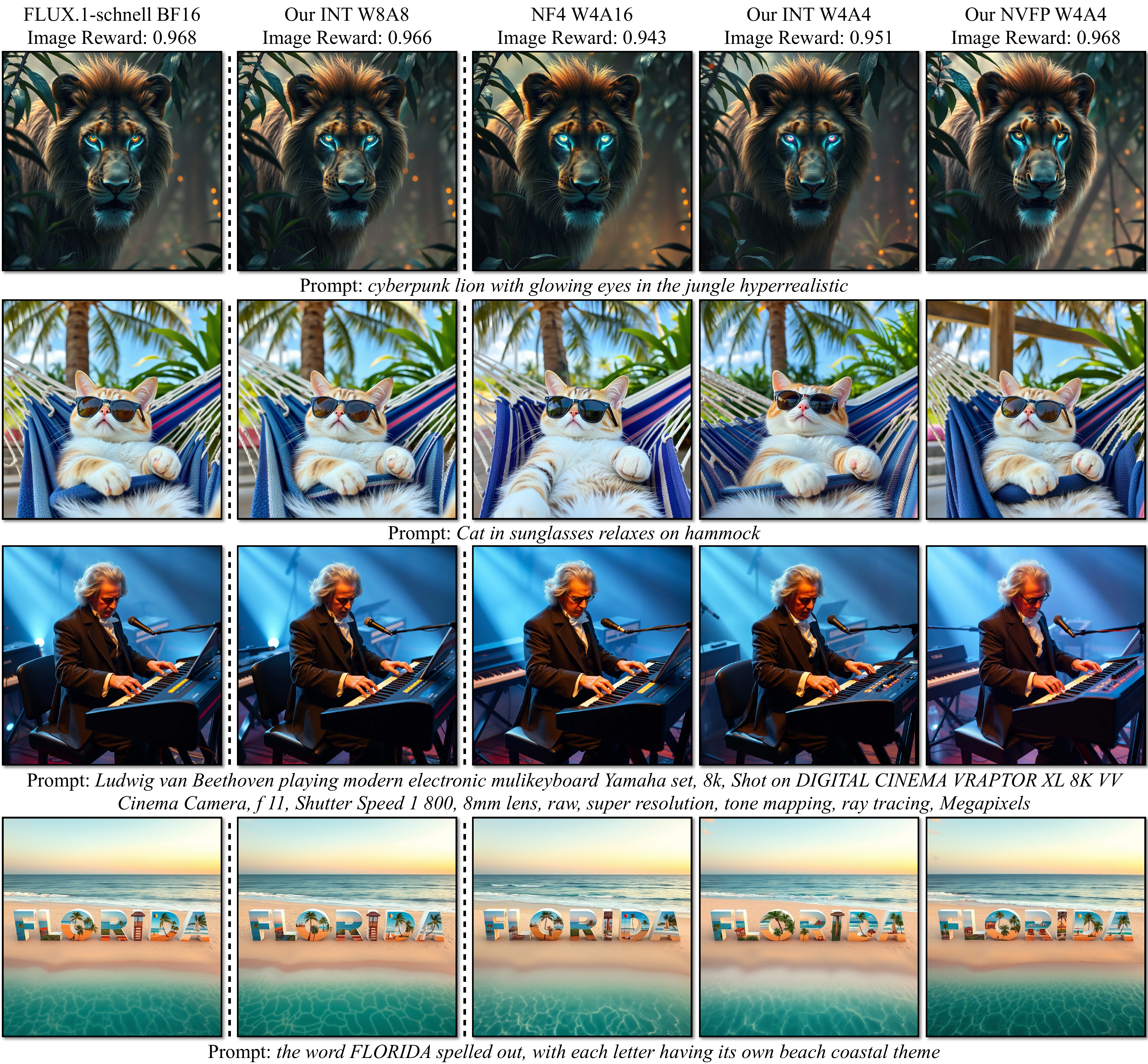}
    \vspace{-5pt}
	\caption{Qualitative visual results of FLUX.1-schnell on MJHQ.}
    \vspace{-15pt}
    \lblfig{flux-schnell-appendix}
\end{figure}
\newpage
\begin{figure}[t]
    \centering
    \includegraphics[width=0.85\linewidth]{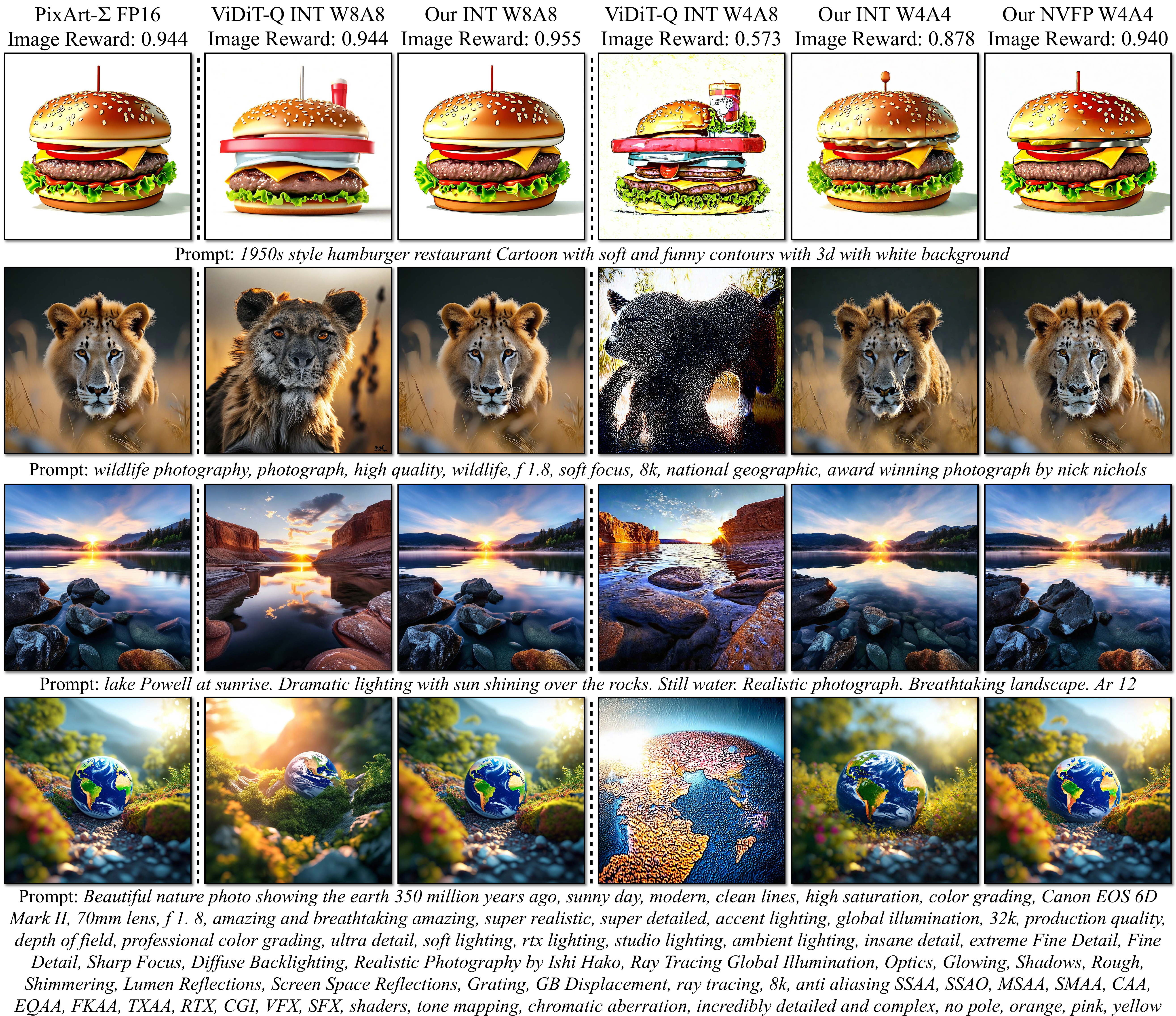}
    \vspace{-5pt}
	\caption{Qualitative visual results of PixArt-$\Sigma$ on MJHQ.}
    \vspace{-20pt}
    \lblfig{pixart-sigma-appendix}
\end{figure}
\begin{figure}[t]
    \centering
    \includegraphics[width=0.85\linewidth]{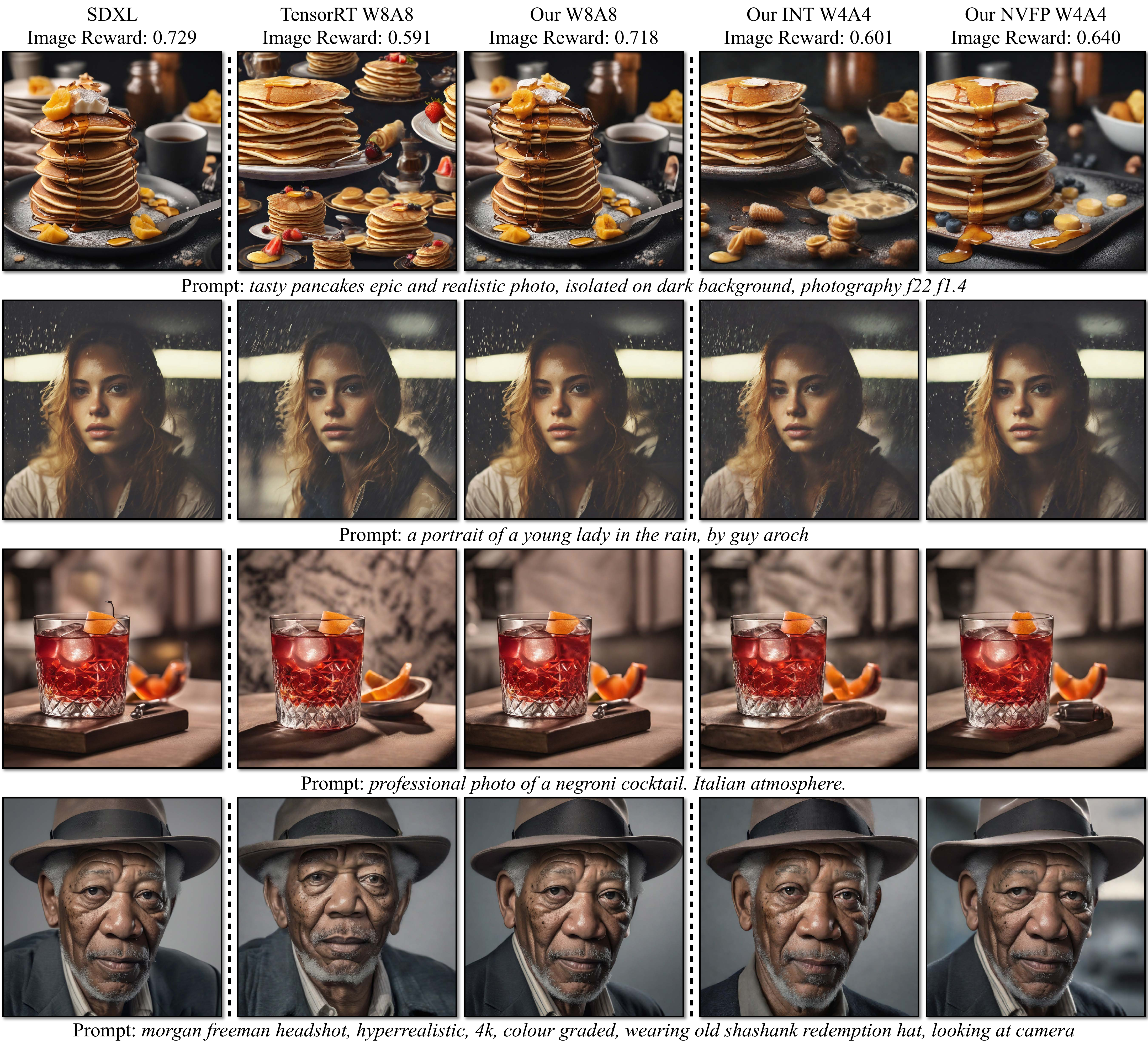}
    \vspace{-5pt}
	\caption{Qualitative visual results of SDXL on MJHQ.}
    \lblfig{sdxl-appendix}
\end{figure}
\begin{figure}[t]
    \centering
    \includegraphics[width=0.85\linewidth]{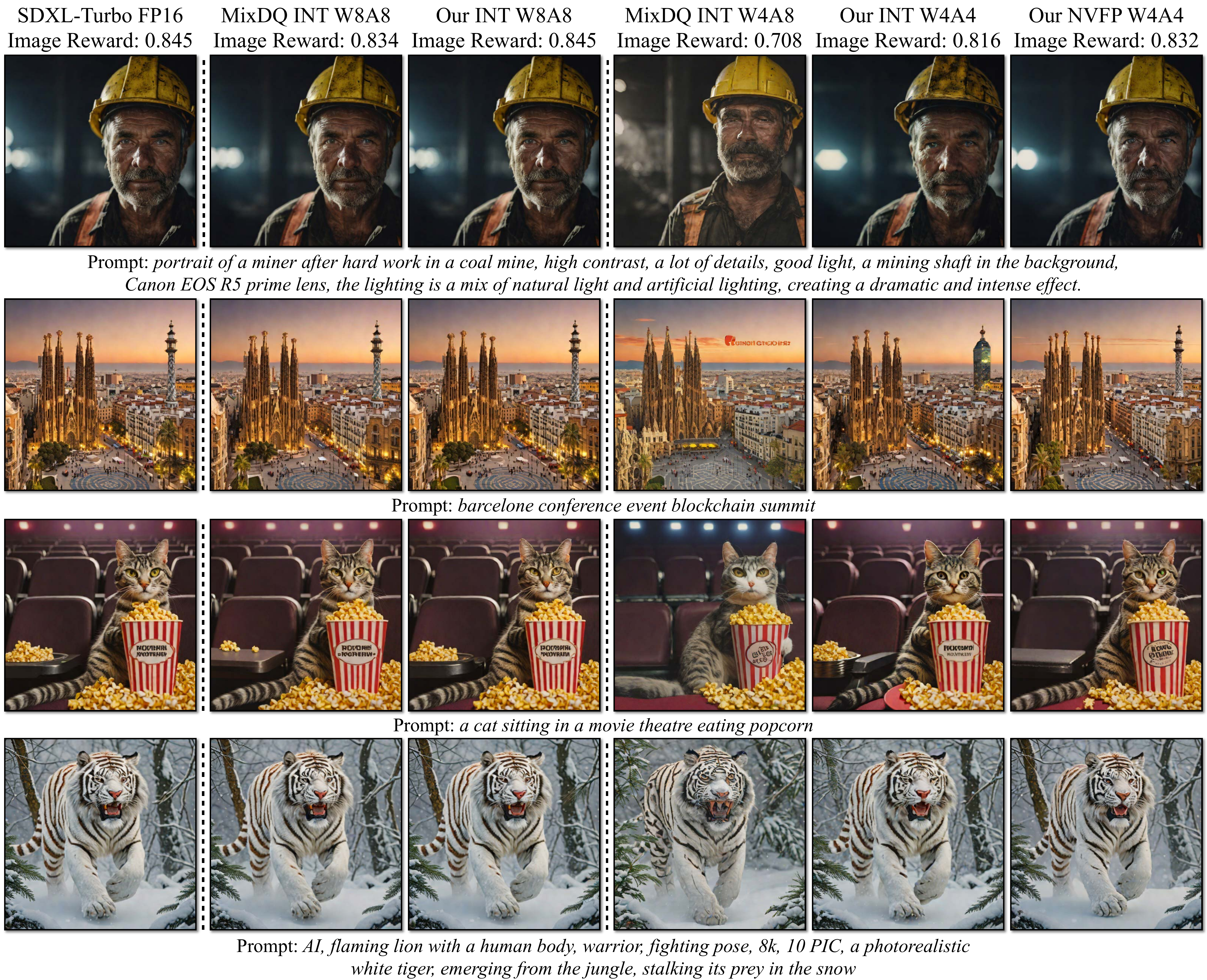}
    \vspace{-5pt}
	\caption{Qualitative visual results of SDXL-Turbo on MJHQ.}
    \vspace{-15pt}
    \lblfig{sdxl-turbo-appendix}
\end{figure}

\clearpage

\subsection{LoRA Results}
\lblapp{LoRA Results}
In \fig{lora-appendix}, we showcase more visual results of applying the aforementioned five community-contributed LoRAs of different styles (\href{https://huggingface.co/XLabs-AI/flux-RealismLora}{Realism}, \href{https://huggingface.co/aleksa-codes/flux-ghibsky-illustration}{Ghibsky Illustration}, \href{https://huggingface.co/alvdansen/sonny-anime-fixed}{Anime}, \href{https://huggingface.co/Shakker-Labs/FLUX.1-dev-LoRA-Children-Simple-Sketch}{Children Sketch}, and \href{https://huggingface.co/linoyts/yarn_art_Flux_LoRA}{Yarn Art}) to our INT4 quantized models.

\begin{figure}[H]
    \centering
    \vspace{-10pt}
    \includegraphics[width=0.89\linewidth]{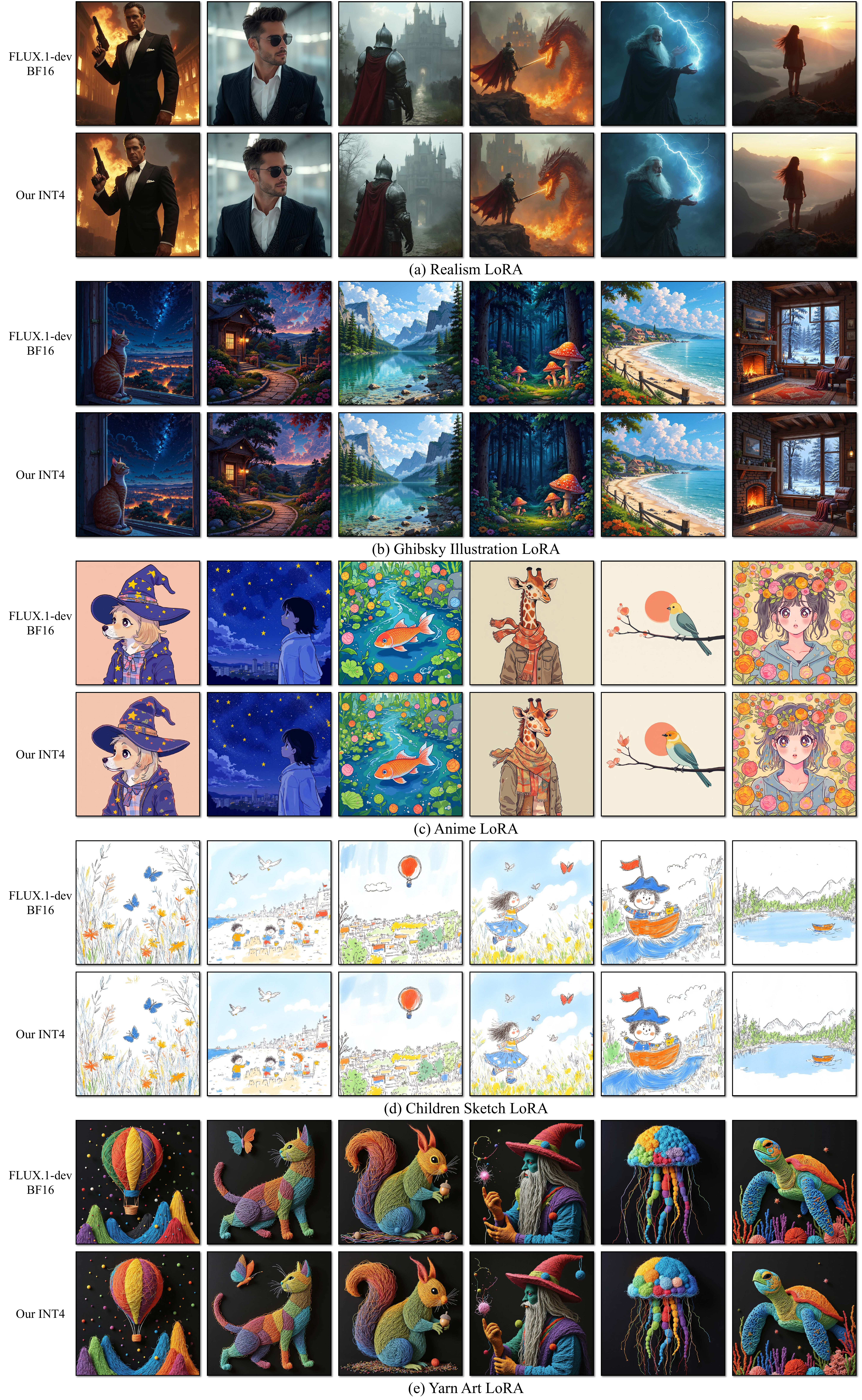}
    \vspace{-5pt}
	\caption{Additional LoRA results on FLUX.1-dev. When applying LoRAs, our INT4 model matches the image quality of the original BF16 model. See \app{Text Prompts} for the detailed used text prompts.}
    \vspace{-15pt}
    \lblfig{lora-appendix}
\end{figure}

\clearpage
\subsection{Additional Ablation of SVDQuant}
\renewcommand \arraystretch{1.}
\begin{table}[h]
    \setlength{\tabcolsep}{4pt}
    \caption{
        Quantitative comparisons of different SVDQuant settings on MJHQ. NVFP4 outperforms INT4. SVDQuant leverages a low-rank branch to ease quantization, significantly enhancing image quality. It can further apply GPTQ to quantize the weight residual, further improving quality.
    }
    \vspace{-5pt}
    \lbltbl{svdquant-ablation}
    \scriptsize \centering
    \begin{tabular}{ccccccc}
    \toprule
    Model & Precision & Low-rank Branch & GPTQ & Image Reward ($\uparrow$) & LPIPS ($\downarrow$) & PSNR ($\uparrow$) \\
    \midrule
        & BF16 & -- & -- & 0.953 & -- & -- \\
        \cmidrule{2-7}
        & \multirow{4}{*}{INT4} & \xmark & \xmark & 0.908 & 0.322 & 18.5\\
        & & \xmark & \cmark & 0.933 & 0.297 & 19.1\\
        & & \cmark & \xmark & 0.926 & 0.256 & 20.1 \\
        FLUX.1-dev& & \cellcolor{mitblue}\cmark & \cellcolor{mitblue}\cmark & \cellcolor{mitblue}\textbf{0.935} & \cellcolor{mitblue}\textbf{0.223} & \cellcolor{mitblue}\textbf{21.0} \\
        \cmidrule{2-7}
        & \multirow{4}{*}{NVFP4} & \xmark & \xmark & 0.928 & 0.244 & 20.3 \\
        & & \xmark & \cmark & 0.936 & \textbf{0.204} & \textbf{21.5} \\
        & & \cmark & \xmark & 0.935 & 0.223 & 20.9 \\
        & & \cellcolor{mitblue}\cmark & \cellcolor{mitblue}\cmark & \cellcolor{mitblue}\textbf{0.937} & \cellcolor{mitblue}0.208 & \cellcolor{mitblue}21.4 \\

        \midrule
        & BF16 & -- & -- & 0.968 & -- & -- \\
        \cmidrule{2-7}
        & \multirow{4}{*}{INT4} & \xmark & \xmark & \textbf{0.962} & 0.345 & 16.3 \\
        & & \xmark & \cmark & \textbf{0.962} & 0.317 & 16.8 \\
        & & \cmark & \xmark & 0.957 & 0.289 & 17.6 \\
        FLUX.1-schnell& & \cellcolor{mitblue}\cmark & \cellcolor{mitblue}\cmark & \cellcolor{mitblue}0.951 & \cellcolor{mitblue}\textbf{0.258} & \cellcolor{mitblue}\textbf{18.3} \\
        \cmidrule{2-7}
        & \multirow{4}{*}{NVFP4} & \xmark & \xmark & 0.957 & 0.280 & 17.5 \\
        & & \xmark & \cmark & 0.956 & 0.247 & 18.5 \\
        & & \cmark & \xmark & \textbf{0.968} & 0.247 & 18.4\\
        & & \cellcolor{mitblue}\cmark & \cellcolor{mitblue}\cmark & \cellcolor{mitblue}\textbf{0.968} & \cellcolor{mitblue}\textbf{0.227} & \cellcolor{mitblue}\textbf{19.0} \\

        \midrule
        & BF16 & -- & -- & 0.944 & -- & -- \\
        \cmidrule{2-7}
        & \multirow{4}{*}{INT4} & \xmark & \xmark & -1.226 & 0.762 & 9.1 \\
        & & \xmark & \cmark & -0.902 & 0.763 & 9.9 \\
        & & \cmark & \xmark & 0.858 & 0.356 & 17.0 \\
        PixArt-$\Sigma$ & & \cellcolor{mitblue}\cmark & \cellcolor{mitblue}\cmark & \cellcolor{mitblue}\textbf{0.878} & \cellcolor{mitblue}\textbf{0.323} & \cellcolor{mitblue}\textbf{17.6}\\
        \cmidrule{2-7}
        & \multirow{4}{*}{NVFP4} & \xmark & \xmark & 0.660 & 0.517 & 14.8 \\
        & & \xmark & \cmark & 0.696 & 0.480 & 15.6 \\
        & & \cmark & \xmark & 0.945 & 0.290 & 18.0\\
        & & \cellcolor{mitblue}\cmark & \cellcolor{mitblue}\cmark &\cellcolor{mitblue}\textbf{0.940} & \cellcolor{mitblue}\textbf{0.271} & \cellcolor{mitblue}\textbf{18.5} \\
        
        \midrule
        & BF16 & -- & -- & 0.952 & -- & -- \\
        \cmidrule{2-7}
        & \multirow{4}{*}{INT4} & \xmark & \xmark & 0.894 & 0.339 & 15.3\\
        & & \xmark & \cmark & 0.881 & 0.288 & 16.4\\
        & & \cmark & \xmark & 0.922 & 0.234 & 17.4 \\
        SANA-1.6B & & \cellcolor{mitblue}\cmark & \cellcolor{mitblue}\cmark & \cellcolor{mitblue}\textbf{0.935} & \cellcolor{mitblue}\textbf{0.220} & \cellcolor{mitblue}\textbf{17.8} \\
        \cmidrule{2-7}
        & \multirow{4}{*}{NVFP4} & \xmark & \xmark & 0.932 & 0.237 & 17.3 \\
        & & \xmark & \cmark & 0.927 & 0.202 & 18.3 \\
        & & \cmark & \xmark & \textbf{0.957} & 0.188 & 18.7 \\
        & & \cellcolor{mitblue}\cmark & \cellcolor{mitblue}\cmark &\cellcolor{mitblue}0.955 & \cellcolor{mitblue}\textbf{0.177} & \cellcolor{mitblue}\textbf{19.0} \\
        
    \bottomrule
    \end{tabular}
\end{table}

In \tbl{svdquant-ablation}, we provide additional quantitative ablation results of SVDQuant on the MJHQ prompt set~\citep{li2024playground}. Across all models, NVFP4 outperforms INT4 due to its native support for smaller microscaling group sizes on Blackwell. \method leverages a low-rank branch to absorb outliers, significantly enhancing image quality in all settings. Additionally, it can incorporate GPTQ~\citep{frantar-gptq} instead of round-to-nearest for weight quantization, further improving quality in most cases. Notably, combining SVDQuant with NVFP4 precision achieves the best results, reaching a PSNR of 21.5 on FLUX.1-dev, closely matching the image quality of the original 16-bit model. In \fig{svdquant-ablation}, we provide qualitative comparisons across different precision settings.

\begin{figure}[h]
    \centering
    \includegraphics[width=\linewidth]{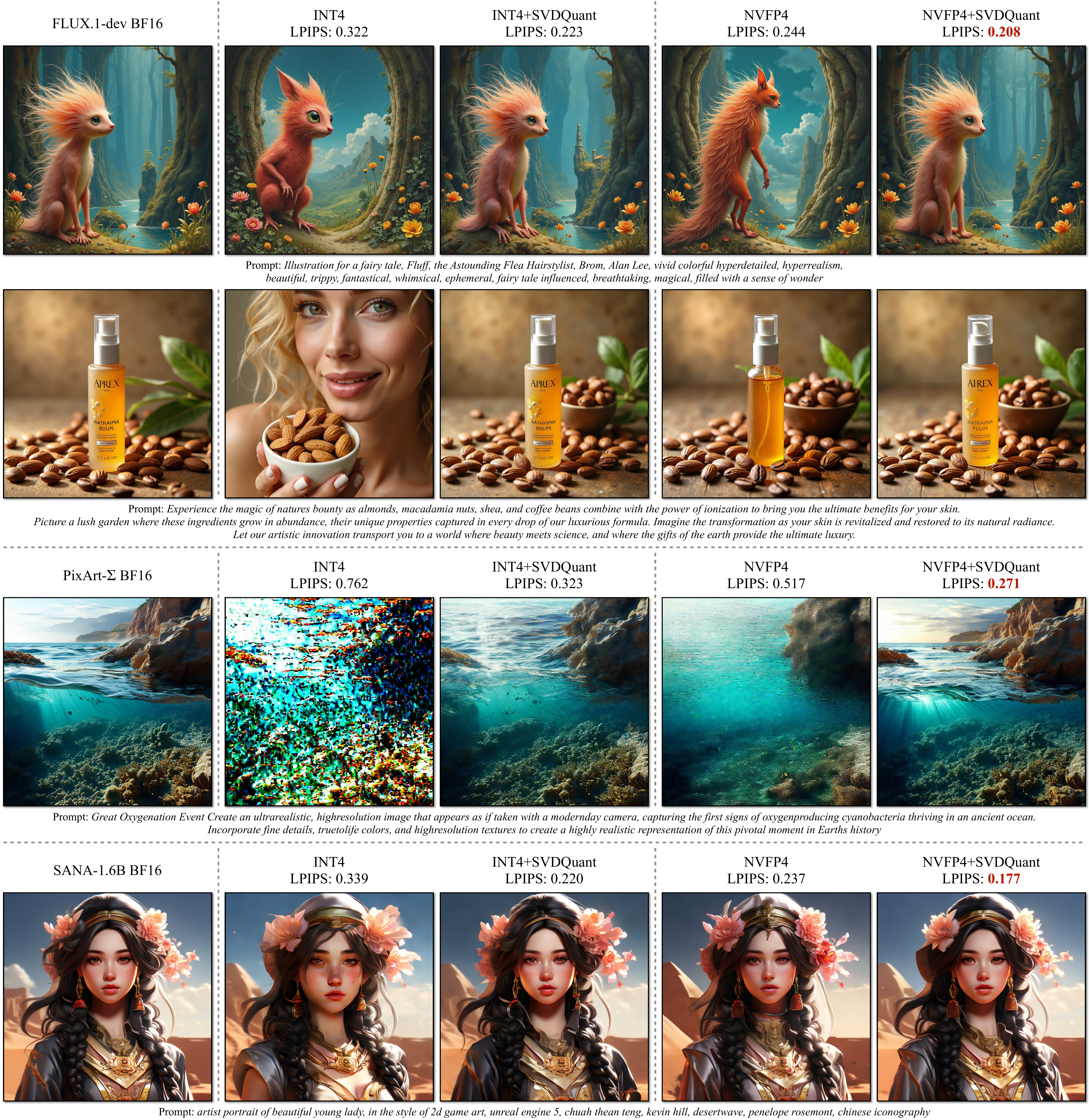}
    \vspace{-15pt}
    \caption{
        Qualitative comparisons of different precisions on MJHQ. NVFP4+SVDQuant yields the highest image fidelity.
    }
    \vspace{-15pt}
    \lblfig{svdquant-ablation}
\end{figure}

\clearpage
\subsection{Latency Results} 
In \tbl{latency-appendix}, we compare FLUX latency on a laptop-level 4090 GPU across different precisions. Compared to INT8, 4-bit quantization delivers a $1.3\times$ speedup. However, without optimization, \method incurs an 18\% overhead due to the low-rank branch. By eliminating redundant memory access, \engine achieves latency comparable to naive INT4.
\renewcommand \arraystretch{1.}
\begin{table}[H]
    \setlength{\tabcolsep}{4pt}
    \caption{
        Single-step latency comparisons of FLUX on a desktop-level 4090 GPU.
    }
    \lbltbl{latency-appendix}
    \scriptsize \centering
    \begin{tabular}{cccccc}
    \toprule
    Method & BF16 & INT8 & \Naive INT4 & \method & \method+\engine \\
    \midrule
    Latency (ms) & 657 & 282 & 212 & 250 & 218\\
    \bottomrule
    \end{tabular}
\end{table}

\subsection{Trade-off of Increasing Rank}
\lblapp{Trade-off of Increasing Rank}
\fig{ablation-rank} presents the results of different rank $r$ in \method on PixArt-$\Sigma$. Increasing the rank from 16 to 64 significantly enhances image quality but increases parameter and latency overhead. In our experiments, we select a rank of 32, which offers a decent quality with minor overhead.

\begin{figure}[h]
    \centering
    \includegraphics[width=\linewidth]{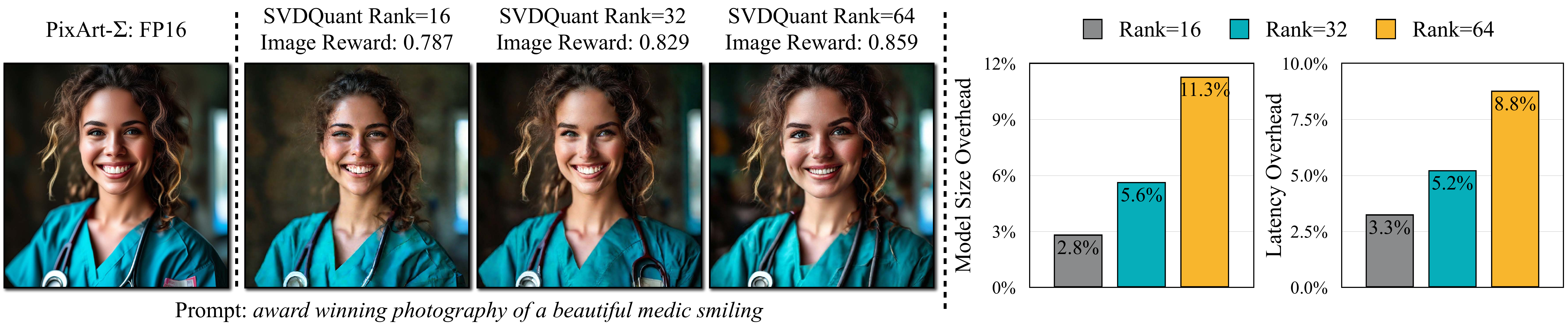}
    \vspace{-15pt}
	\caption{
        Increasing the rank $r$ of the low-rank branch in SVDQuant can enhance image quality, but it also leads to higher parameter and latency overhead.
    }
    \vspace{-15pt}
    \lblfig{ablation-rank}
\end{figure}

\subsection{Trade-off between Quality and Bitwidth} We evaluate LPIPS across different bitwidths for various quantization methods on PixArt-$\Sigma$ and FLUX.1-schnell using the MJHQ dataset in \fig{tradeoff-appendix}, with weights and activations sharing the same bitwidth. Following the convention~\citep{xiao2023smoothquant,lin2024awq,lin2025qserve,li2023q,zhao2024atom,dettmers2022gpt3}, for bitwidths above 4, we apply per-channel quantization; for 4 or below, we use per-group quantization (group size 64). SVDQuant consistently outperforms naive quantization and SmoothQuant. Notably, on PixArt--$\Sigma$ and FLUX.1-schnell, our 4-bit results match 7-bit and 6-bit naive quantization, respectively.

Our SVDQuant can still generate images in the 3-bit settings on both PixArt-$\Sigma$ and FLUX.1-schnell, performing much better than SmoothQuant. Below this precision (e.g., W2A4 or W4A2), SVDQuant cannot produce images either since 2-bit symmetric quantization is essentially a ternary quantization. Prior work~\citep{ma2024era,wang2023bitnet} has shown that ternary neural networks require quantization-aware training even for weight-only quantization to adapt the weights and activations to the low-bit distribution.
\begin{figure}[H]
    \centering
    \vspace{-10pt}
    \includegraphics[width=0.7\linewidth]{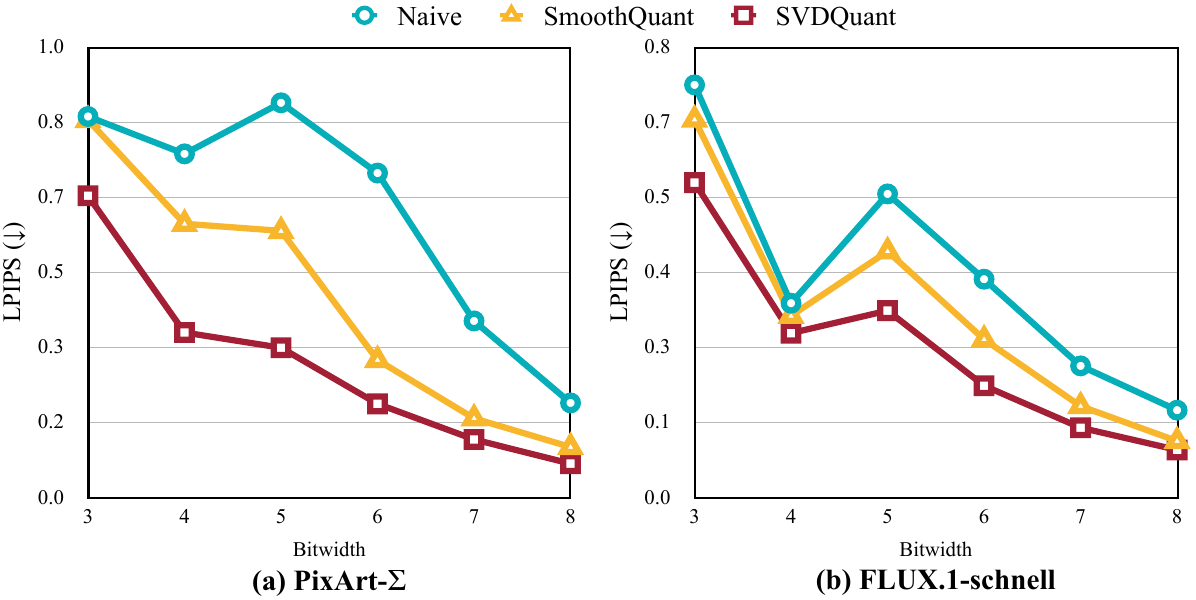}
	\caption{LPIPS of different quantization methods on PixArt-$\Sigma$ and FLUX.1-schnell across different bitwidths.}
    \vspace{-15pt}
    \lblfig{tradeoff-appendix}
\end{figure}

\section{Text Prompts}
\lblapp{Text Prompts}
Below we provide the text prompts we use in \fig{lora} (from left to right).
\begin{Verbatim}[breaklines=true, fontsize=\footnotesize]
a man in armor with a beard and a sword
GHIBSKY style, a fisherman casting a line into a peaceful village lake surrounded by quaint cottages
girl, neck tuft, white hair, sheep horns, blue eyes, nm22 style
sketched style, A squirrel wearing glasses and reading a tiny book under an oak tree
a panda playing in the snow, yarn art style
\end{Verbatim}
The text prompts we use in \fig{lora-appendix} are (in the rasterizing order):
\begin{Verbatim}[breaklines=true, fontsize=\footnotesize]
A male secret agent in a tuxedo, holding a gun, standing in front of a burning building
A handsome man in a suit, 25 years old, cool, futuristic
A knight in shining armor, standing in front of a castle under siege
A knight fighting a fire-breathing dragon in front of a medieval castle, flames and smoke
A male wizard with a long white beard casting a lightning spell in the middle of a storm
A young woman with long flowing hair, standing on a mountain peak at dawn, overlooking a misty valley

GHIBSKY style, a cat on a windowsill gazing out at a starry night sky and distant city lights
GHIBSKY style, a quiet garden at twilight, with blooming flowers and the soft glow of lanterns lighting up the path
GHIBSKY style, a serene mountain lake with crystal-clear water, surrounded by towering pine trees and rocky cliffs
GHIBSKY style, an enchanted forest at night, with glowing mushrooms and fireflies lighting up the underbrush
GHIBSKY style, a peaceful beach town with colorful houses lining the shore and a calm ocean stretching out into the horizon
GHIBSKY style, a cozy living room with a view of a snow-covered forest, the fireplace crackling and a blanket draped over a comfy chair

a dog wearing a wizard hat, nm22 anime style
a girl looking at the stars, nm22 anime style
a fish swimming in a pond, nm22 style
a giraffe with a long scarf, nm22 style
a bird sitting on a branch, nm22 minimalist style
a girl wearing a flower crown, nm22 style

sketched style, A garden full of colorful butterflies and blooming flowers with a gentle breeze blowing
sketched style, A beach scene with kids building sandcastles and seagulls flying overhead
sketched style, A hot air balloon drifting peacefully over a patchwork of fields and forests below
sketched style, A sunny meadow with a girl in a flowy dress chasing butterflies
sketched style, A little boy dressed as a pirate, steering a toy ship on a small stream
sketched style, A small boat floating on a peaceful lake, surrounded by trees and mountains


a hot air balloon flying over mountains, yarn art style
a cat chasing a butterfly, yarn art style
a squirrel collecting acorns, yarn art style
a wizard casting a spell, yarn art style
a jellyfish floating in the ocean, yarn art style
a sea turtle swimming through a coral reef, yarn art style
\end{Verbatim}

\end{document}